\documentclass{article}
\usepackage{times}
\usepackage{import}
\usepackage{natbib}
\usepackage{url}
\usepackage{fancyvrb}
\usepackage{subfigure}
\usepackage{comment}
\usepackage{mdframed}
\usepackage{enumitem}
\usepackage{relsize}
\usepackage{framed}
\usepackage{graphicx}
\usepackage{wrapfig}

\usepackage[capitalize]{cleveref}  

\crefname{nlem}{Lemma}{Lemmas}
\crefname{nprop}{Proposition}{Propositions}
\crefname{ncor}{Corollary}{Corollaries}
\crefname{nthm}{Theorem}{Theorems}
\crefname{assumption}{Assumption}{Assumptions}

\usepackage{jhh-misc2}

\usepackage[accepted]{icml2015}

\newcommand{\distSTP}{\distNamed{STP}}
\newcommand{\GGP}{$\Gamma\Gamma\mathrm{P}$\xspace}
\newcommand{\HGGP}{H\GGP}

\suppresscomments
\boldshortcuts

\allowdisplaybreaks[3]

\begin{document}

\icmltitlerunning{JUMP-Means: SVA for MJPs}

\twocolumn[

\icmltitle{JUMP-Means: Small-Variance Asymptotics for \\ Markov Jump Processes}

\icmlauthor{Jonathan H.~Huggins*}{jhuggins@mit.edu}
\icmlauthor{Karthik Narasimhan*}{karthikn@csail.mit.edu}
\icmlauthor{Ardavan Saeedi*}{ardavans@mit.edu}
\icmlauthor{Vikash K.~Mansinghka}{vkm@mit.edu}
\icmladdress{Computer Science and Artificial Intelligence Laboratory,
MIT \\
*These authors contributed equally and are listed alphabetically.}

\icmlkeywords{small-variance asymptotics, Markov jump processes, Bayesian nonparametric models}

\vskip 0.3in
]

\newcommand{\fix}{\marginpar{FIX}}
\newcommand{\new}{\marginpar{NEW}}
\setlength{\marginparwidth}{20mm}

\begin{abstract}
Markov jump processes (MJPs) are used to model a wide range of 
phenomena from disease progression to RNA path folding. 
However, maximum likelihood estimation of parametric models leads to degenerate
trajectories and inferential performance is poor in nonparametric models.
We take a small-variance asymptotics (SVA) approach 
to overcome these limitations. 
We derive the small-variance asymptotics for parametric
and nonparametric MJPs for both
directly observed and hidden state models.
In the parametric case we obtain a novel objective
function which
leads to non-degenerate trajectories. 
To derive the nonparametric version we introduce the
gamma-gamma process, a novel extension to the 
gamma-exponential process. 
We propose algorithms for each of these 
formulations, which we call \emph{JUMP-means}.
Our experiments demonstrate that JUMP-means is competitive with
or outperforms widely used MJP inference approaches 
in terms of both speed and reconstruction accuracy. 
\end{abstract}

\begin{figure*}[t!]
\begin{minipage}{0.48\textwidth}
\includegraphics[trim = 0mm 0mm 0mm 0mm, clip,width=\textwidth]{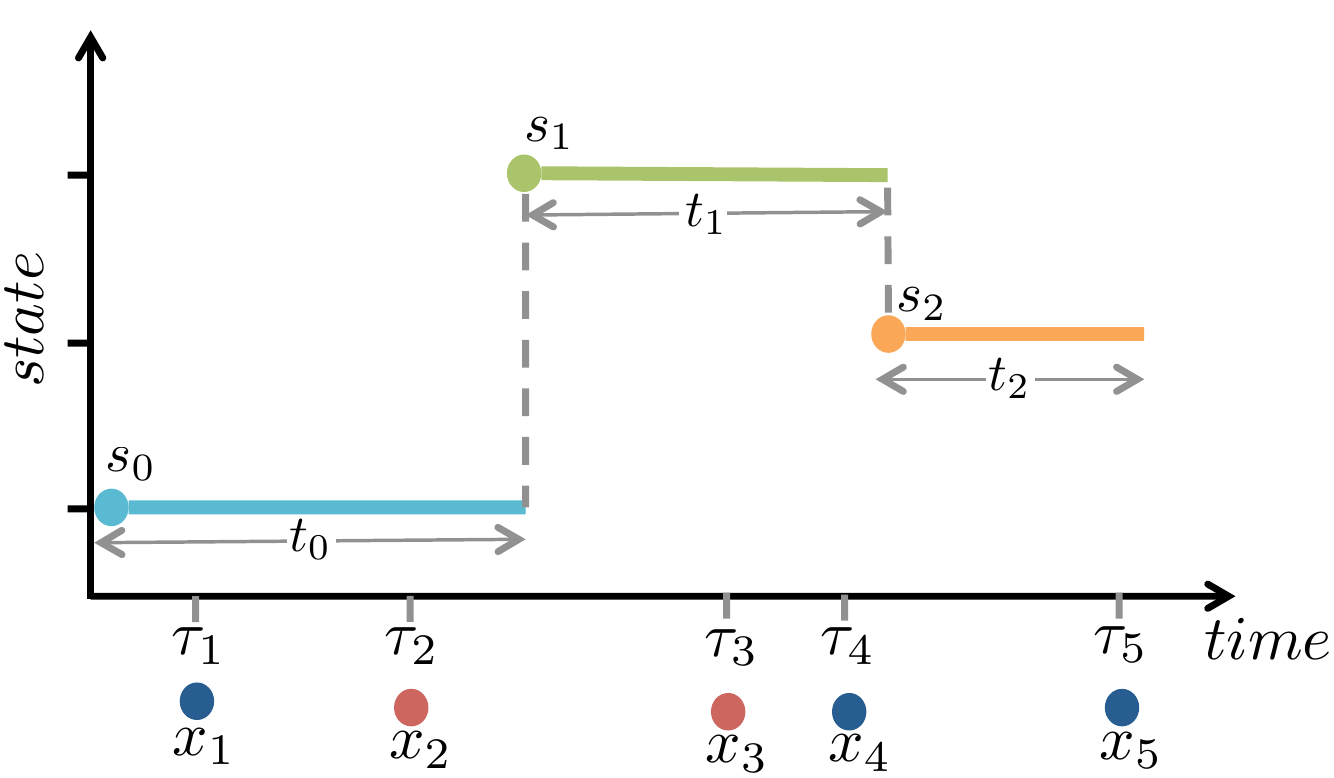}
\end{minipage}
~
\begin{minipage}{0.47\textwidth}
\begin{tabular}{p{\textwidth}}
\hline
\textbf{Notation} \\
\hline
$M$: number of states\\
$\pi$:  initial state distribution\\
$P$:  state transition matrix, with entries $p_{ss'}$\\
$\lambda_{s}$:  transition rate for state $s$\\
$\mcU = (s_{0}, t_{0}, s_{1}, {t}_1, \dots, s_{K-1}, t_{K-1}, s_{K})$: MJP trajectory \\
$\mcS$: the states corresponding to $\mcU$ \\
$\mcT$: the times corresponding to $\mcU$ \\
$\mcO = \{(\tilde t_{i},\tilde s_{i})\}$: observation times and states of
DOMJP* \\
$\boldsymbol\tau = (\tau_{1},\dots, \tau_{L})$: observation times of HMJP\\
$\mcX = (x_{1},\dots,x_{L})$: observations of HMJP\\
$\rho_{sn}$: probability of observing $x_{\ell} = n$ when in state $s$\\
\hline
\end{tabular}
\end{minipage}
\caption{\textbf{Left:} Illustrative example for an HMJP (Section \ref{sec:parametric-hidden}) 
with three hidden states ($M=3$) and two possible observation values ($N=2$).  
The observations $\mcX$, their times $\boldsymbol\tau$, an (arbitrary) sample 
MJP trajectory $\mcU = (s_{0}, t_{0}, s_{1}, {t}_1, s_{2}, {t}_2)$. 
\textbf{Right:} Notation used for parametric MJPs.  *DOMJP~= directly observed MJP. }
\label{fig:illustration}
\vspace{-3mm}
\end{figure*}

\section{Introduction}

Markov jump processes (MJPs) are continuous-time, discrete-state 
Markov processes in which state durations are exponentially 
distributed according to state-specific rate parameters.
A stochastic matrix controls the probability of transitioning
between pairs of states. 
MJPs have been used to construct probabilistic models
either when the state of a system is observed directly,
such as with disease progression~\citep{mandel2010estimating} and
RNA path folding~\citep{Hajiaghayi:2014}, 
or when the state is only observed indirectly, as in 
corporate bond rating~\citep{bladt2009efficient}. 
For example, consider the important clinical task of 
analyzing physiological
signals of a patient in order to detect abnormalities. 
Such signals include heart rate, blood pressure, respiration, 
and blood oxygen level.
For an ICU patient, an abnormal state might be 
the precursor to a cardiac arrest event while for an epileptic, the 
state might presage a seizure \citep{PhysioNet}. 
How can the latent state of the patient be inferred by a Bayesian 
modeler, so that, for example, an attending nurse can be notified when
a patient enters an abnormal state? 
MJPs offer one attractive approach to
analyzing such physiological signals.

Applying an MJP model to physiological signals presents a
challenge: the number of states is unknown and 
must be inferred using, for example, Bayesian nonparametric methods.
However, efficient inference in nonparametric MJP models is 
a challenging problem, where existing methods based on
particle MCMC scale poorly and mix slowly~\citep{saeedi2011priors}. 
Current optimization-based methods such as expectation maximization (EM)
are inapplicable if the state size is countably infinite; hence, 
they cannot be applied to Bayesian nonparametric MJP models, as
we would like to do for physiological signals. 

Furthermore, although MJPs are viewed as more realistic than 
their discrete-time counterparts in many fields~\citep{Rao:2013}, 
degenerate solutions for the maximum 
likelihood (ML) trajectories for both directly and indirectly observed
cases~\citep{perkins2009maximum}, and non-existence of 
the ML transition matrix (obtained from EM) for some indirectly 
observed cases~\citep{bladt2009efficient} present inferential challenges. 
Degenerate ML trajectories occur when some of the jump times are infinitesimal, 
which severely undermines the practicality of such approaches. For instance,
a trajectory which predicts a patient's seizure for an infinitesimal amount of time
is of limited use to the medical staff.
Fig.~\ref{fig:npb} shows an example of the degeneracy problem. 

In this paper, we take a \emph{small-variance asymptotics} (SVA) approach to
develop an optimization-based framework for efficiently estimating the most probable trajectories (states) for 
both parametric and nonparametric MJP-based models.
Small-variance asymptotics has recently proven to be 
useful in estimating the parameters and inferring the latent states in 
rich probabilistic models.  
SVA extends the well-known connection between mixtures of 
Gaussians and $k$-means: as the variances of the Gaussians
approach zero, the \emph{maximum a posteriori} solution to the
mixture of Gaussians model degenerates to $k$-means solution~\citep{Kulis:2012}. 
The same idea can be applied to obtain well-motivated objective 
functions that correspond to a latent variable model for which 
scalable inference via standard methods like MCMC is challenging.
SVA has been applied to (hierarchical) Dirichlet process mixture 
models~\citep{Kulis:2012,jiang2012small},  
Bayesian nonparametric latent feature models~\citep{Broderick:2013}, hidden Markov models (HMMs),
and infinite-state HMMs~\citep{Roychowdhury:2013}.

We apply the SVA approach to both parametric and Bayesian 
nonparametric MJP models to 
obtain what we call the \emph{JUMP-means} objective functions. 
In the parametric case, we derive a novel objective function 
which does not suffer from maximum likelihood's solution degeneracy,
leading to more stable and robust inference procedures 
in both the directly observed and hidden state cases. 
Infinite-state MJPs (iMJPs) are constructed from the hierarchical 
gamma-exponential process (H$\Gamma$EP)~\citep{saeedi2011priors}.
In order to apply SVA to iMJPs, we generalize the H$\Gamma$EP
to obtain the first deterministic procedure (we
know of) for inference in Bayesian nonparametric MJPs. 

We evaluate JUMP-means on several synthetic and real-world datasets 
in both the parametric and Bayesian non-parametric cases. 
JUMP-means performs on par with or better than existing methods,
offering an attractive speed-accuracy tradeoff. 
We obtain significant improvements in the non-parametric case, 
gaining up to a 20\% reduction in mean error on the task of observation
reconstruction. 
In summary, the JUMP-means approach leads to algorithms that 1)
are applicable to MJPs with Bayesian nonparametric priors;
2) provide non-degenerate solutions for the most probable trajectories;
and 3) are comparable to or outperform other standard methods of inference
both in terms of speed and reconstruction accuracy.

\section{Background}
\label{sec:background}

\subsection{Markov Jump Processes}

A \emph{Markov jump process} (MJP) is defined by 
(a) a finite (or countable) state space, which we identify with 
the integers $[M] \defined \theset{1,\dots,M}$;
(b) an initial state probability distribution $\pi$;
(c) a (stochastic) state transition matrix $P$ with
$p_{ss} = 0$ for all $s \in [M]$; and
(d) a state dwell-time rate vector
$\boldsymbol\lambda \defined (\lambda_{1},\dots, \lambda_{M})$. 
The process begins in a state $s_{0} \dist \pi$. 
When the process enters a state $s$, it remains
there for a dwell time that is exponentially 
distributed with parameter $\lambda_{s}$. 
When the system leaves state $s$, it transitions
to state $s' \ne s$ with probability $p_{ss'}$. 

A \emph{trajectory} of the MJP is a sequence of states and 
a dwell time for each state, except for the final state:
$\mcU \defined \mcU_{T} \defined (s_{0},t_{0}, s_{1},t_{1},\dots, s_{K-1}, t_{K-1}, s_{K})$.
Implicitly, $K$ (and thus $\mcU$) is a random variable such that $t_{K-1} < T$
and the system is in state $s_{K}$ at time $T$. 
Let $\mcS \defined \mcS_{T} \defined (s_{0}, s_{1},\dots, s_{K})$
and $\mcT \defined \mcT_{T} \defined (t_{0}, t_{1},\dots, t_{K-1})$ be 
the sequences of states and times corresponding to $\mcU$. 
The probability of a trajectory is given by
\[
p(\mcU \given \pi, P, \boldsymbol\lambda)
&= \ind[t_{\cdot} \le T] e^{-\lambda_{s_{K}}(T - t_{\cdot})}\pi_{s_0}\label{eq:trajectory-probability} \\ 
&\phantom{=~}\times \textstyle\prod^{K}_{k = 1} \lambda_{s_{k-1}} e^{-\lambda_{s_{k-1}}t_{k-1}}p_{s_{k-1}s_{k}}, \nonumber
\]
where $t_{\cdot} \defined \sum^{K-1}_{k=0} t_k$
and $\ind[\cdot]$ is the indicator function. 
In many cases when the states are directly observed,
the initial state and the final state are observed,
in which case it is straightforward to obtain
a likelihood from \eqref{eq:trajectory-probability}. 

A hidden state MJP (HMJP) is an MJP in which 
the states are observed indirectly according 
to a likelihood model $p(x \given s)$, $s \in [M]$,
$x \in X$, where $X$ is some observation space. 
The times of the observations 
$\boldsymbol\tau = (\tau_{1}, \dots, \tau_{L})$
are chosen independent of $\mcU$, so the 
probability of the observations 
$\mcX \defined (x_{1},\dots,x_{L})$
is given by
$
p(\mcX \given \mcU,\boldsymbol\tau) 
= \prod_{\ell=1}^{L} p(x_{\ell} \given s_{\tau_{\ell}}),
$
where, with an abuse of notation, we write $s_{\tau}$ for
the state of the MJP at time $\tau$. 

\subsection{Previous Approaches to MJP Inference}
\label{sec:previous_approaches}
There are a number of existing approaches to inference 
and learning in MJPs. An expectation-maximization (EM) algorithm
can be derived, but it cannot be applied to models with countably infinite states, so it is not suitable for iMJPs~\citep{lange2014latent} (iMJPs are detailed in Section \ref{sec:nonparametric}). 
Moreover, with discretely observed data, the maximum-likelihood
estimate with finite entries for the transition matrix obtained from EM
may not exist~\citep{bladt2005statistical}. 

Maximum likelihood inference amounts to finding 
$\max_{\mcU}\ln p(\mcU \given \pi, P, \boldsymbol\lambda)$,
which can be carried out efficiently using dynamic
programming~\citep{perkins2009maximum}. 
However, maximum likelihood solutions for the trajectory
are degenerate: only an infinitesimal amount 
of time is spent in each state, except for the state
visited with the smallest rate parameter (i.e., longest 
expected dwell time). 
Such a solution is unsatisfying and unintuitive
because the dwell times are far from their expected values. 
Thus, maximum likelihood inference produces 
results that are unrepresentative of the 
model behavior. 

Markov chain Monte Carlo methods have also been
developed, but these can be slow and their convergence 
is often difficult to diagnose~\citep{Rao:2013}. 
Recently, a more efficient Monte Carlo method was 
proposed in~\citet{Hajiaghayi:2014} which is based 
on particle MCMC (PMCMC) methods~\citep{Andrieu:2010}.
This approach addresses the issue of efficiency, but 
since it marginalizes over the jump points, it cannot 
provide probable trajectories.

\subsection{Small-variance Asymptotics}

Consider a Bayesian model $p(D \given Z, \theta, \sigma^{2})p(Z, \theta)$
in which the likelihood terms contain a variance parameter $\sigma^{2}$. 
Given some data $D$, a point estimate for the parameters $\theta$ and latent variables
$Z$ of the model can obtained by maximizing the posterior
$p(Z, \theta \given D, \sigma^{2}) \propto p(D \given Z, \theta, \sigma)p(Z, \theta)$,
resulting in a \emph{maximum a posteriori} (MAP) estimate.
In the SVA approach~\citep{Broderick:2013}, the MAP optimization is considered in the limit
as the likelihood variance parameter is taken to zero: $\sigma^{2} \to 0$.
Typically, the small-variance limit leads to a much simpler optimization
than the MAP optimization with non-zero variance. 
For example, the MAP objective for a Gaussian mixture model simplifies to 
the $k$-means objective. 

\section{Parametric MJPs}
\label{sec:parametric}

\subsection{Directly Observed MJP}
\label{sec:parametric-direct}

Consider the task of inferring likely state/dwell-time 
sequences given $\mcO = \{ (\tilde t_{i}, \tilde s_{i}) \}_{i=1}^{I}$,
the times at which the system was directly observed and the states of the system
at those times. 
For simplicity we assume that $\tilde t_{0} = 0$ and
that all times are in the interval $[0,T]$.
Let $s(\mcU, t)$ be the state of the system following trajectory
$\mcU$ at time $t$. 
The likelihood of a sequence is
\[
\begin{split}
\lefteqn{\ell(\mcU \given \mcO, P, \boldsymbol\lambda) = \ind[t_{\cdot} \le T]\prod_{i=1}^{I} \ind[s(\mcU, \tilde t_{i}) = \tilde s_{i}]} \\
&\times \left(\prod^{K}_{k = 1} \lambda_{s_{k-1}} e^{-\lambda_{s_{k-1}}t_{k-1}}p_{s_{k-1}s_{k}}\right)e^{-\lambda_{s_{K}}(T - t_{\cdot})} 
\label{eq:likelihood-observe-ends}
\end{split}
\]
We also place a gamma prior on the rate parameters 
$\boldsymbol\lambda$ (detailed below).
Instead of relying on MAP estimation, we apply a small variance 
asymptotics analysis to obtain a more stable objective function.
Following~\citep{jiang2012small}, we scale the distributions 
by an inverse variance parameter $\beta$ and then maximize the scaled 
likelihood and prior in the limit $\beta \to \infty$ (i.e., as the variance 
goes to zero). 

Scaling the exponential distribution 
$f(t;\lambda)=\lambda\exp(-\lambda t)$ produces 
the two-parameter family with
\(
\lefteqn{\ln f(t;\lambda,\beta) =} \\
& -\beta\left( 
      \lambda t - \ln t  - \ln \lambda 
    - \frac{\beta \ln \beta - \ln \Gamma(\beta)}{\beta} 
    + \frac{\ln t}{\beta}  \right),
\)
which is the density of a
gamma distribution with shape parameter $\beta$ and rate 
parameter $\beta\lambda$.
Hence, the mean of the scaled distribution is $\frac{1}{\lambda}$ and its variance is $\frac{1}{\lambda^{2}\beta}$. 
Letting $F(t; \lambda, \beta)$ denote the CDF corresponding to $f(t; \lambda,\beta)$, 
we have $1-F(t; \lambda,\beta) = \frac{\Gamma(\beta, \beta \lambda t)}{\Gamma(\beta)}$,
where $\Gamma(\cdot,\cdot)$ is the upper incomplete gamma function. 

The multinomial distribution is scaled by 
the parameter $\hat{\beta} \defined \xi \beta$.
Writing the likelihood with the scaled exponential 
families (and dropping indicator variables) yields:
\[
\lefteqn{\ell(\mcU| \mcO, P, \boldsymbol\lambda)} \nonumber \\
&\propto \exp \left\{
  - \beta \left(\frac{\ln \Gamma(\beta) 
  - \ln\Gamma(\beta, \beta \lambda_{s_{K}}t_{\cdot})}{\beta} \right. \right. \\
&\phantom{\propto~\exp\{~}  +  \sum^{K-1}_{k=0}\left(\xi \ln p_{s_{k}s_{k+1}} + \lambda_{s_{k}} t_k - \ln \lambda_{s_{k}}t_k \right)\nonumber \\
&\phantom{\propto~\exp\{~} \left. \left.  
   + \sum^{K-1}_{k=0}\left(- \frac{\beta \ln \beta - \ln \Gamma(\beta)}{\beta} + \frac{\ln t_k}{\beta} \right)\right)\right\}. \nonumber
\]
The modified likelihood is for a jump process which is no longer Markov when $\beta \ne 1$.
We also place a $\distGam(\alpha_{\lambda}, \alpha_{\lambda}\mu_{\lambda})$ 
prior on each $\lambda_{i}$ and set $\alpha_{\lambda} = \xi_{\lambda}\beta$.
It can be shown (see the Supplementary Material for details) that, when $\beta \to \infty$, 
the MAP estimation problem becomes 
\[
\min_{\mcU, \boldsymbol\lambda, P} \bigg\{ 
&   \xi \sum^{K-1}_{k=0} \ln p_{s_{k}s_{k+1}} + \sum^{K-1}_{k=0}\left(\lambda_{s_k} t_k - \ln \lambda_{s_k} t_k - 1\right) \nonumber \\
& + \ind[\lambda_{s_{K}}t_{\cdot} \geq 1](\lambda_{s_{K}}t_{\cdot} - \ln \lambda_{s_{K}}t_{\cdot} - 1) \label{eq:mjp-objective} \\
& + \xi_{\lambda}\sum_{s=1}^{M}(\mu_{\lambda}\lambda_{s} - \ln \lambda_{s} - 1)  \bigg\}. \nonumber
\]
The optimization problem \eqref{eq:mjp-objective} is 
very natural and offers far greater stability than 
maximum likelihood optimization. 
As with maximum likelihood, 
the $\ln p_{s_{k}s_{k+1}}$ terms penalize transitions with small probability. 
The term $h(t_{k}) \defined \lambda_{s_k} t_k - \ln \lambda_{s_k} t_k - 1$ is convex
and minimized when $t_{k} = 1/\lambda_{s_{k}}$, the expected value 
of the dwell time for state $s_{k}$. 
As $t_{k} \to 0$, $h(t_{k})$ approaches $\infty$, while for 
$t_{k} \gg 1/\lambda_{s_{k}}$, $h(t_{k})$ grows approximately linearly.
Thus, times very close to zero are heavily penalized while
times close to the expected dwell time are penalized
very little. 
The term  
$\ind[\lambda_{s_{K}}t_{\cdot} \geq 1](\lambda_{s_{K}}t_{\cdot} - \ln \lambda_{s_{K}}t_{\cdot} - 1)$
penalizes the time $t_{\cdot}$ spent in state $s_{k}$ 
{so far} in the same manner
as a regular dwell time when $t_{\cdot}$ is \emph{greater} than the 
expected value of the dwell-time. 
However, when $t_{\cdot}$ is \emph{less} than the expected value
there is no cost, which is quite natural since the system
may remain in state $s_{k}$ for longer than $t_{\cdot}$ 
--- i.e., there should not be a large penalty for $t_{\cdot}$ 
being less than its expected value. 
Finally, parameters $\xi_{\lambda}$ and $\mu_{\lambda}$ have a
very natural interpretation (cf.~\eqref{eq:lambda-update} below):
they correspond to \emph{a priori} having $\xi_{\lambda}$  
dwell times of length $\mu_{\lambda}$ for each state. 

\textbf{Comparison to Maximum Likelihood}
MJP trajectories estimated using maximum likelihood (MLE) are usually trivial, with the system
spending almost all its time in a single state (with the smallest $\lambda$), with infinitesimal dwell 
times for the other states. This poor behavior of MLE is due to the fact that the mode of 
$\distExp(\lambda)$, which is favored by the MLE, is 0,
even though the mean is $1/\lambda$.\footnote{Note that placing priors on 
the rate parameters, as we do, does not affect the degeneracy of the ML trajectory.}
The SVA optimization, on the other hand, does 
give trajectories that are representative of the true behavior 
because the SVA terms of the form $\lambda t - \ln(\lambda t) - 1$ 
are optimized at $1/\lambda$ (i.e., at the mean of $\distExp(\lambda)$).
We demonstrate the superior behavior of the SVA in the concrete example 
of estimating disease progression in patients in Section \ref{sec:experiments}.

\subsection{Hidden State MJP}
\label{sec:parametric-hidden}

For an HMJP, the likelihood of a valid trajectory is
\[
\begin{split}
&p(\mcU \given \mcX, \boldsymbol\tau, P, \boldsymbol\lambda) 
= \left(\prod_{\ell=1}^{L} p(x_{\ell} \given s_{\tau_{\ell}})\right) \\
&\times \left(\prod^{K}_{k = 1} \lambda_{s_{k-1}} e^{-\lambda_{s_{k-1}}t_{k-1}}p_{s_{k-1}s_{k}}\right)e^{-\lambda_{s_{K}}(T - t_{\cdot})}. \label{eq:likelihood-hidden-mjp}
\end{split}
\]
Hence, the only difference between the directly observed 
case and the HMJP is the addition of the observation likelihood 
terms. 
Because multinomial observations are commonly used in
MJP applications, that is the case we consider here. 
Let $N$ denote the number of possible observations
and $\rho_{sn}$ be the probability of observing
$x_{\ell} = n$ when $s_{\tau_{\ell}} = s$.
The observation likelihoods are scaled in the 
same manner as the transition probabilities, 
but with $\hat{\beta} = \zeta\beta$.
Thus, for the HMJP, we obtain:
\[
\min_{\mcU, \boldsymbol\lambda, P, \rho} \bigg\{
&   \zeta\sum_{\ell=1}^{L} \ln\rho_{s_{\tau_{\ell}}x_{\ell}}
  + \xi \sum^{K-1}_{k=0}\ln p_{s_{k}s_{k+1}} \nonumber \\
&  + \sum^{K-1}_{k=0}\left(\lambda_{s_k} t_k - \ln \lambda_{s_k} t_k - 1\right) \label{eq:hmjp-objective} \\
& + \ind[\lambda_{s_{K}}t_{\cdot} \geq 1](\lambda_{s_{K}}t_{\cdot} - \ln \lambda_{s_{K}}t_{\cdot} - 1) \nonumber \\
& + \xi_{\lambda}\sum_{s=1}^{M}(\mu_{\lambda}\lambda_{s} - \ln \lambda_{s} - 1) \bigg\}. \nonumber
\]

\subsection{Algorithm}
\label{sec:parametric_alg}

Optimizing the JUMP-means objectives in \eqref{eq:mjp-objective} and
\eqref{eq:hmjp-objective} is non-trivial due to the fact 
that we do not know the number of jumps in the MJP, and 
the combinatorial explosion in the sequences with the number
of jump points. 
The terms involving the continuous variables $t_k$ (dwell times)
present an additional complexity.

We therefore resort to an alternating minimization procedure 
to optimize the JUMP-means objective function, similar in 
spirit to the one used in \citet{Roychowdhury:2013}. 
In each iteration of the optimization process, we first use a 
modified Viterbi algorithm to obtain the most likely state sequence. 
Then, we use convex optimization to distribute the jump points 
optimally with respect to the values from $\boldsymbol\lambda$ for
the current state sequence.

\textbf{Directly Observed MJP}
When optimizing \eqref{eq:mjp-objective}, there may be 
many sequences ($\mcO$'s) available, representing
distinct realizations of the process. 
We use the following algorithm to optimize \eqref{eq:mjp-objective}:
\vspace{-1mm}
\begin{enumerate}[label=\arabic*., leftmargin=*]
\itemsep0em
\item Initialize the state transition matrix $P$ and rate vector 
$\boldsymbol\lambda$ with uniform values. 

\item For every observation sequence $\mcO$, instantiate the jump points 
by adding one jump point between every pair of observations, 
in addition to the start and end points.

\item For each $\mcO$, use a modified Viterbi algorithm.
to find the best state sequence to optimize \eqref{eq:mjp-objective}, 
while keeping the jump points fixed. 
The modified algorithm includes
the dwell time penalty terms, which are dependent upon the 
assignment of states to the time points.

\item Optimize the dwell times $t_k$ with the state sequences of the trajectories fixed.

\item Optimize $P$ and $\boldsymbol\lambda$ with the other variables fixed.
The optimal values can be obtained in closed form. 
For example, if there is only a single observation sequence $\mcO$ with 
corresponding inferred trajectory $\mcS$, then 
\[
p_{mj} &= \frac{n_{mj}}{\sum_{j=1}^{M} n_{mj}}, \quad m,j \in [M] \\
\lambda_m &= \frac{\xi_{\lambda} + \sum_k \ind[s_{k}=m]}{\xi_{\lambda}\mu_{\lambda} + \sum_k \ind[s_{k}=m] t_k}, \label{eq:lambda-update}
\]
where $n_{mj}$ denotes to the number of transitions from state 
$m$ to state $j$ in $\mcS$.

\item Repeat steps 3-5 until convergence.
\end{enumerate}

\textbf{Beam Search Variant}
We note that the optimization procedure just described 
is restrictive since the number of jump points is fixed and
the jump points are constrained by the observation 
boundaries. 
To eliminate this, we also tested a beam search variant 
of the algorithm to allow for the creation and removal
of jump points, but found it did not have much impact
in our experiments.

\textbf{Hidden State MJP}
The algorithm to optimize the hidden
state MJP JUMP-means objective \eqref{eq:hmjp-objective}
is similar to that for optimizing \eqref{eq:mjp-objective},
but with three modifications. 
First, in place of $\mcO$, we have the indirect observations of the 
states $\mcX$. 
Second, observation likelihood terms containing $\boldsymbol\rho$ are included in the objective 
minimized by the Viterbi optimization (step 3).
Finally, an additional update is performed in step 5 for
each of the observation distributions $\boldsymbol\rho_{m}$: 
\[
\rho_{mn} &= 
\frac{\sum_\ell \ind[s_{\tau_{\ell}}=m] \ind[x_{\ell}=n] }{\sum_\ell \ind[s_{\tau_{\ell}}=m]}  \label{eq:rho_update}
\]
for $m \in [M]$ and $n \in [N]$. 
If each $\boldsymbol\rho_{m} \defined (\rho_{m1},\dots,\rho_{mN})$ is initialized to be uniform,
then the algorithm converges to a poor local minimum,
so we add a small amount of random noise 
to each uniform $\boldsymbol\rho_{m}$.

\begin{table*}[t]
\vspace{-5mm}
\caption{Statistics and Mean observation reconstruction error for the various models on different datasets. 
\textbf{Key:} BL~=~Baseline; P~=~parametric; SVA~=~JUMP-means; 
NP~=~nonparametric; DO~=~directly observed; H~=~hidden; 
MS~=~multiple sclerosis data set; MIMIC~=~blood pressure data set. 
*Best result obtained by running EM with various number of hidden states (up to 12).
}
\vspace{-1mm}
\begin{center}
\begin{tabular}{lcccccccc} 
\multicolumn{4}{c}{Data Set} & \multicolumn{5}{|c}{Mean Error (\%)} \\
 & {\# Points} & {Held Out} & \# States & \multicolumn{1}{|c}{BL} & {EM} & {MCMC} & {SVA} & {PMCMC}  \\
\hline
Synthetic 1 (P-DO) & 10,000 & 30 \%  & 10 & 69.7 &  \textbf{40.2} & \textbf{41.9} & \textbf{41.2}  & -   \\
Synthetic 2 (P-H) & 10,000 & 30 \%  & 5 & 51.8 &  \textbf{42.9} & 74.6 & 46.5  & - \\
MS (P-DO) & 390 & 50 \%  &  3 & 51.2  & \textbf{26.2} & 48.1 & \textbf{25.4}  & - \\
MIMIC (NP-H) & 2,208 & 25 \%  & - & 42.3 & 25.7* & - & \textbf{24.3}   & 30.9  \\
\end{tabular}
\end{center}
\vspace{-4mm}
\label{tbl:experiments}
\end{table*}

\section{Bayesian Nonparametric MJPs}
\label{sec:nonparametric}

We now consider the Bayesian nonparametric MJP (iMJP) model.
The iMJP is based on the hierarchical gamma-exponential
process (H$\Gamma$EP), which is constructed from the gamma-exponential 
process ($\Gamma$EP). 
We denote the Moran gamma process with base measure $H$ and rate 
parameter $\gamma$ by $\distGamP(H, \gamma)$~\citep{Kingman:1993}.
The H$\Gamma$EP generates a state/dwell-time sequence
$s_{0}, t_1,s_1, t_2, s_2, t_{3}, s_{3}, \dots$
(with $s_{0}$ assumed known) according 
to~\citep{saeedi2011priors}: 
\[
\mu_{0} &\dist \distGamP(\alpha_{0}H_{0}, \gamma_{0}), \\
\mu_{i} &\given \mu_{0} \distiid \distGamP(\mu_{0}, \gamma), \quad i = 1,2,\dots, \\
s_{k} &\given \{\mu_{i}\}_{i=0}^{\infty}, \mcU_{k-1} \dist \bar\mu_{s_{k-1}}, \\
t_{k} &\given  \{\mu_{i}\}_{i=0}^{\infty}, \mcU_{k-1} \dist \distExp(\|\mu_{s_{k-1}}\|), \label{eq:hgep-time-dist}
\]
where $H_{0}$ is the base probability measure, $\alpha_{0}$ is 
a concentration parameter, 
$\mcU_{k} \defined (s_{0}, t_1,s_1,  t_2, s_2, \dots, t_{k-1}, s_{k})$,
$\bar\mu_{i} \defined \mu_{i}/\|\mu_{i}\|$, and $\|\mu\|$ denotes
the total mass of the measure $\mu$. 
As in the parametric case, we must replace the exponential
distribution in \eqref{eq:hgep-time-dist} with the scaled exponential distribution.
After an appropriate scaling of the rest of the hyperparameters,
we obtain the hierarchical gamma-gamma process (\HGGP).
The definition and properties of the \HGGP are given in the 
Supplementary Material.

Let $M$ denote the number of used states, $K_{m}$ the 
number of transitions out of state $m$, and $\mu_{ij}$ the
mass on the $j$-th component of the measure $\mu_{i}$. 
For $0 \le i \le M, 1 \le j \le M$, let 
$\bar\pi_{ij} \defined \bar\mu_{ij}$ and for
$0 \le i \le M$, let 
$\bar\pi_{i,M+1} \defined 1 - \sum_{j=1}^{M} \bar\mu_{ij}$. 
Let $\bt^{*}_{m} \defined (t^{*}_{m1},\dots,t^{*}_{mK_{m}})$ be the 
waiting times following state $m$ and define $t^{*}_{m\cdot} \defined \sum_{j=1}^{K_{m}}t^{*}_{mj}$.
In order to retain the effects of the hyperparameters in the 
asymptotics, set $\alpha_{0} = \exp(-\xi_{1}\beta)$ and
$\gamma_{0} = \kappa_{0} = \xi_{2}$. 
It can then be shown that (see the Supplementary Material for details),
when $\beta \to \infty$, the iMJP MAP estimation problem becomes
\[
&\min_{K,\mcU_{K},\bar\pi,\boldsymbol\rho}\phantom{+}~\zeta\sum_{\ell=1}^{L} \ln\rho_{s_{\tau_{\ell}}x_{\ell}} 
+ \xi \sum_{k=1}^{K} \ln\bar\pi_{s_{k-1}s_{k}} \nonumber \\
&\phantom{\min~} + \xi_{1}M + \sum_{m=1}^{M} \xi_{2}\kl{\bar\pi_{0}}{\bar\pi_{m}}  \label{eq:likelihood-infinite-mjp} \\
&\phantom{\min~} - \sum_{m=1}^{M} \left\{\textstyle\sum_{j=1}^{K_{m}} \ln t^{*}_{mj} - K_{m}\ln \left([\gamma + t^{*}_{m\cdot}]/K_{m}\right) \right\}. \nonumber
\]
Like its parametric counterpart, the Bayesian nonparametric 
cost function penalizes dwell durations very close to zero
via the $\ln t^{*}_{mj}$ terms. 
In addition, 
there are penalties for the number of states and the 
state transitions. 
The observation likelihood term in \eqref{eq:likelihood-infinite-mjp} 
favors the creation of new states to minimize the JUMP-means objective, while the 
state penalty $\xi_{1}M$ and the non-linear penalty term 
$K_{m}\ln\left([\gamma + t^{*}_{m\cdot}]/K_{m}\right)$ counteracts 
the formation of a long tail of states with very few data points.
The $\gamma$ hyperparameter introduces an additional,
nonlinear cost for each additional state --- if a state
is occupied for $\Omega(\gamma)$ time, then the
$\gamma$ term for that state does not have much effect
on the cost. 
The KL divergence terms between $\bar\pi_{0}$ and $\bar\pi_{m}$
arise from the hierarchical structure of the prior, biasing 
the transition probabilities $\bar\pi_{m}$ to be similar to the  
prior $\bar\pi_{0}$.

\subsection{Algorithm}

For the iMJP case, we have the extra 
variables $M$ and $\{ \bar\pi_m \}_{m=0}^{M}$ to optimize. 
In addition, the number of variables to optimize
depends on the number of states in our model. 
The major change in the algorithm from the parametric case is 
that we must propose and then accept or reject the addition of new states.
We propose the following algorithm for optimizing the iMJP:
\vspace{-2mm}
\begin{enumerate}[label=(\arabic*), leftmargin=*]
\itemsep0em
\item Initialize $\boldsymbol\rho$, $\bar\pi_0$ and $\bar\pi_1$ with uniform values and set the number of states $M = 1$. 
\item For each observation sequence, apply the Viterbi algorithm and 
update the times using the new objective function in 
\eqref{eq:likelihood-infinite-mjp}, analogously to steps (3) and (4) 
in the parametric algorithm.
\item Perform MAP updates for $\boldsymbol\rho$ (as in \eqref{eq:rho_update}) and $\bar\pi$:
\[
\bar\pi_{mj} &= \frac{\xi n_{mj} + \xi_{2} \bar\pi_{0j}}{\xi \sum_{j=1}^{M} n_{mj} + \xi_{2} \bar\pi_{0j}}, \quad m,j \in [M]  \\
\bar\pi_{0j} &\propto \prod_{m=1}^{M} \bar\pi_{mj}^{1/M}, \quad j \in [M].
\]
\vspace{-6mm}
\item For every state pair $m, m' \in [M]$, form a new state $M+1$ by considering all transitions from $m$ to $m'$ and reassigning all observations $x_{\ell}$ that were assigned to $m'$ to the new state.
Update $\bar\pi$ and $\boldsymbol\rho$  to estimate the overall objective function for every new set of $M+1$ states formed in this way and accept the state set that minimizes the objective. If no such set exists, do not create a new state and revert back to the old $\bar\pi$ and $\boldsymbol\rho$.
\item Repeat steps 2-4 until convergence.
\end{enumerate}

\begin{figure*}[t]
\vspace{-2mm}
\centering
\subfigure[]{%
	\includegraphics[trim = 0mm 0mm 20mm 0mm, clip, scale = 0.22]{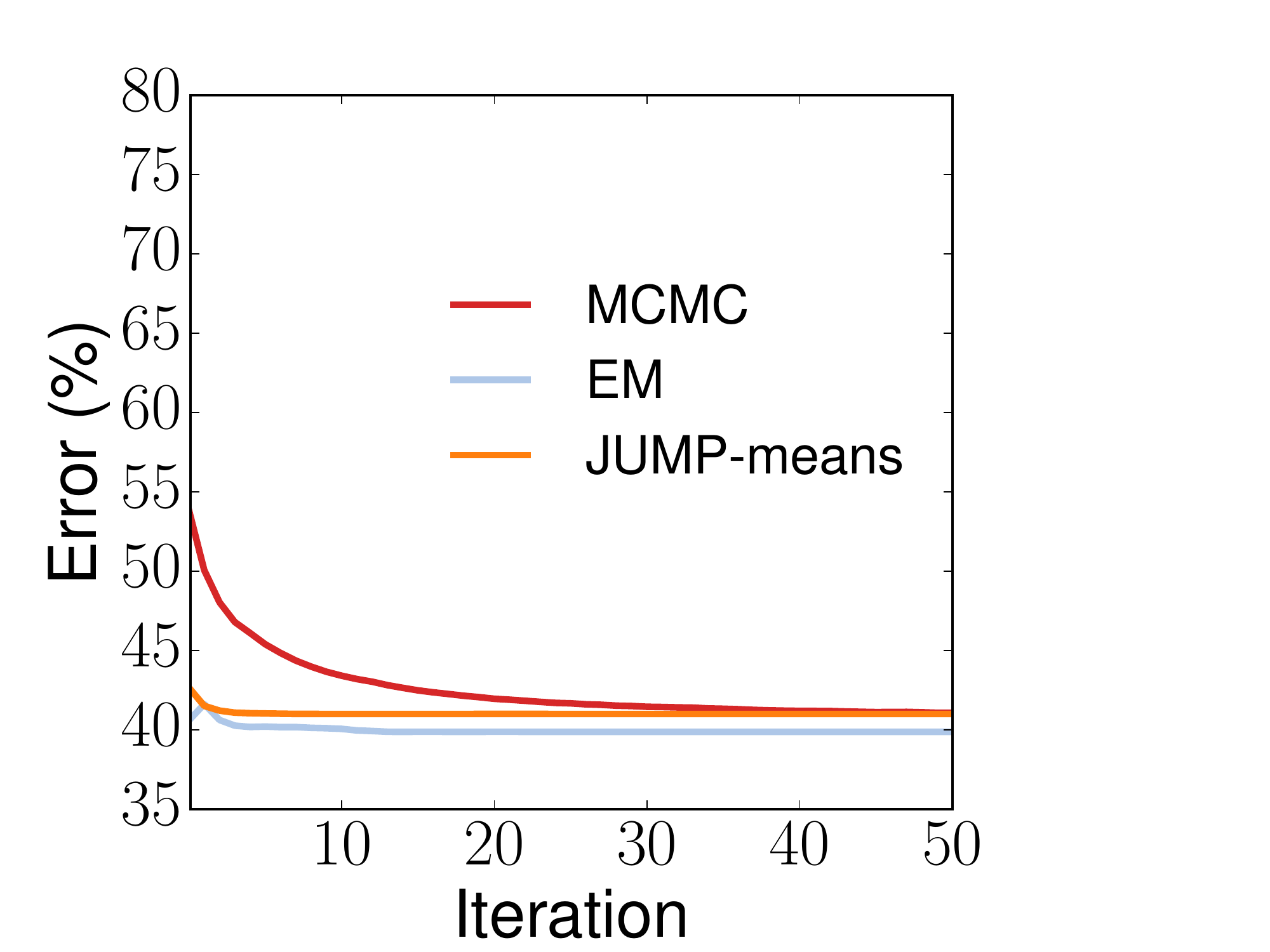}%
}
\subfigure[]{%
	\includegraphics[trim = 0mm 0mm 20mm 0mm, clip, scale = 0.22]{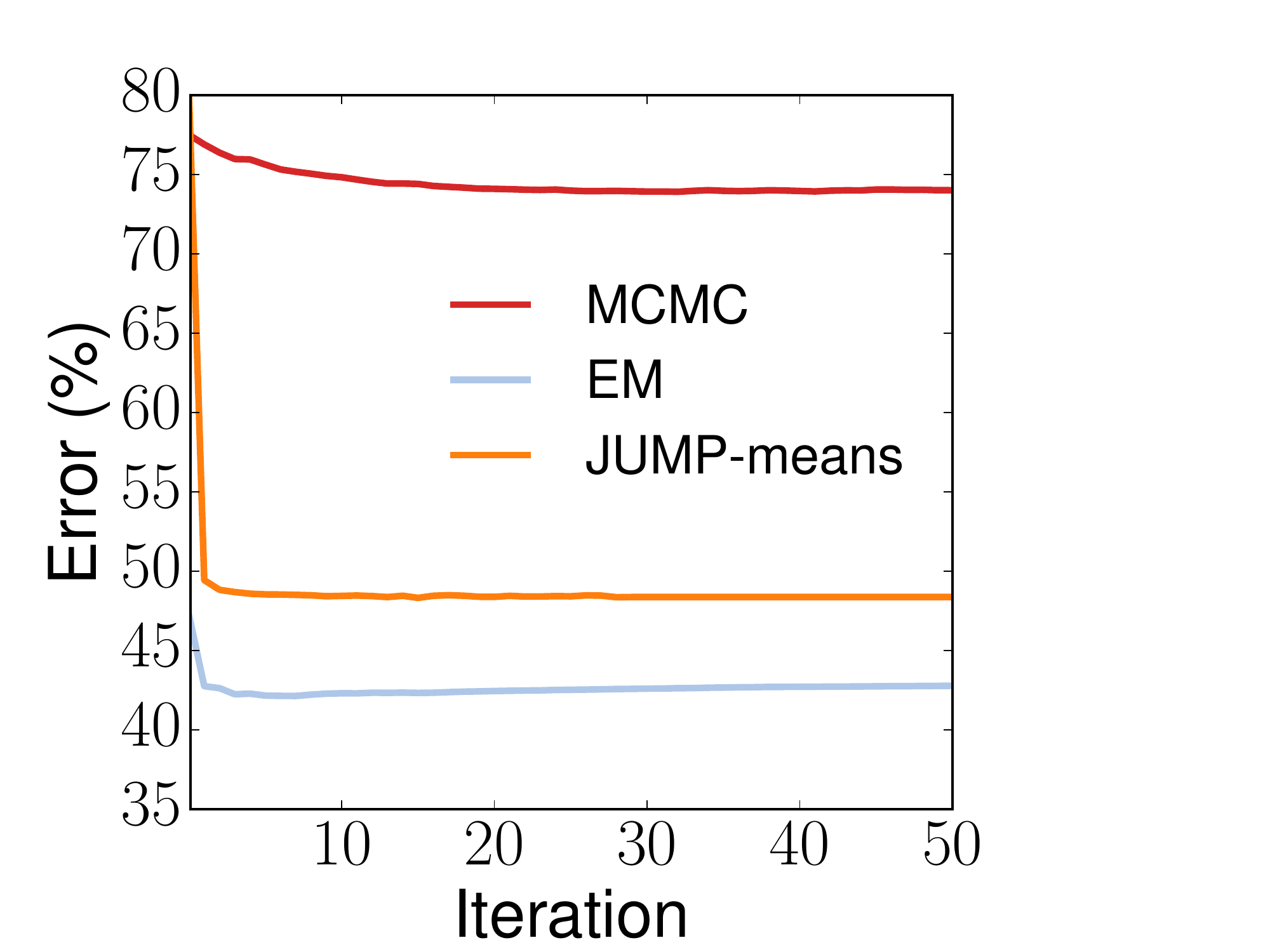}%
}
\subfigure[]{%
	\includegraphics[trim = 0mm 0mm 20mm 0mm, clip, scale = 0.22]{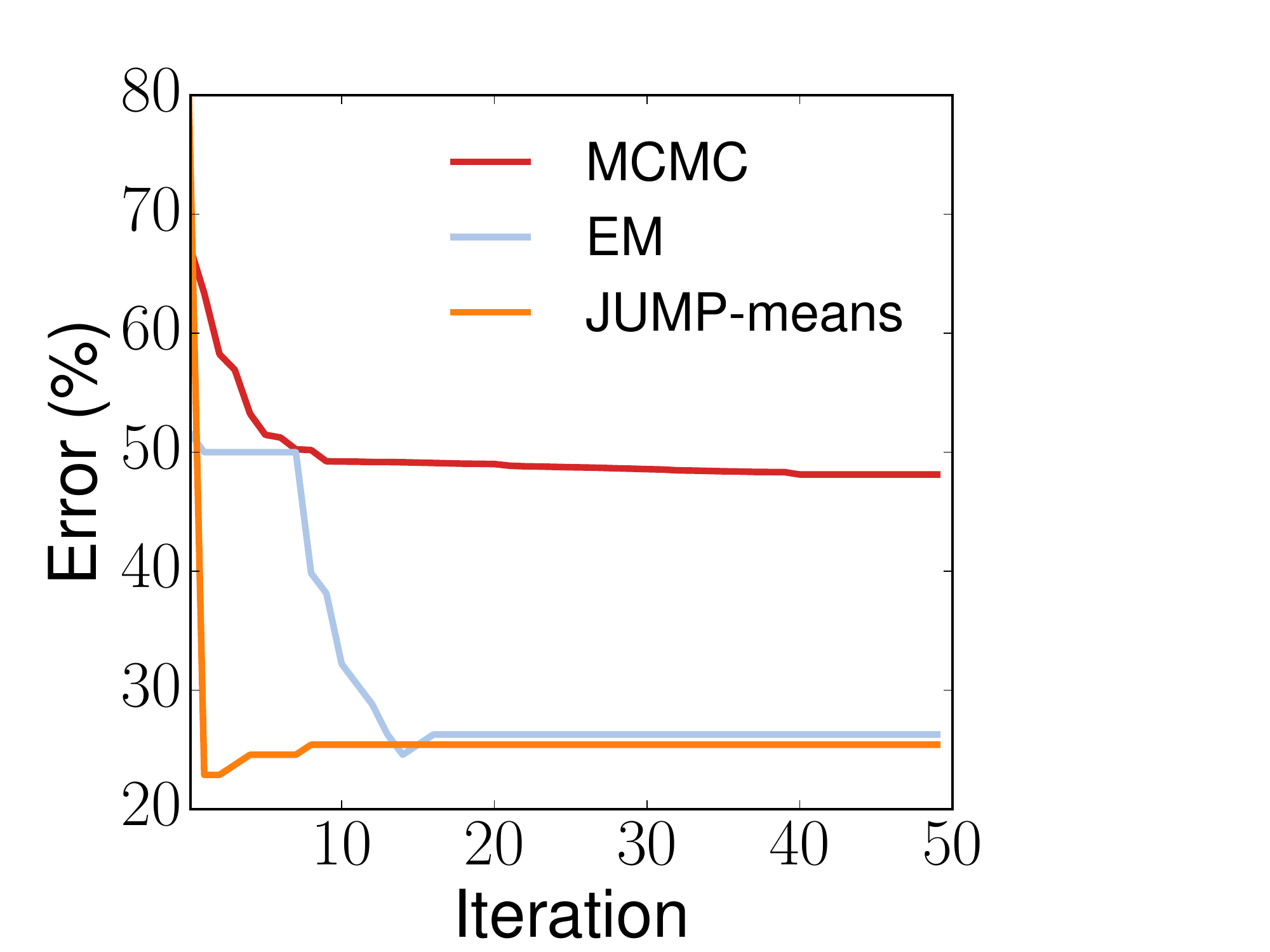}%
}
\subfigure[]{%
	\includegraphics[trim = 0mm 0mm 20mm 0mm, clip, scale = 0.22]{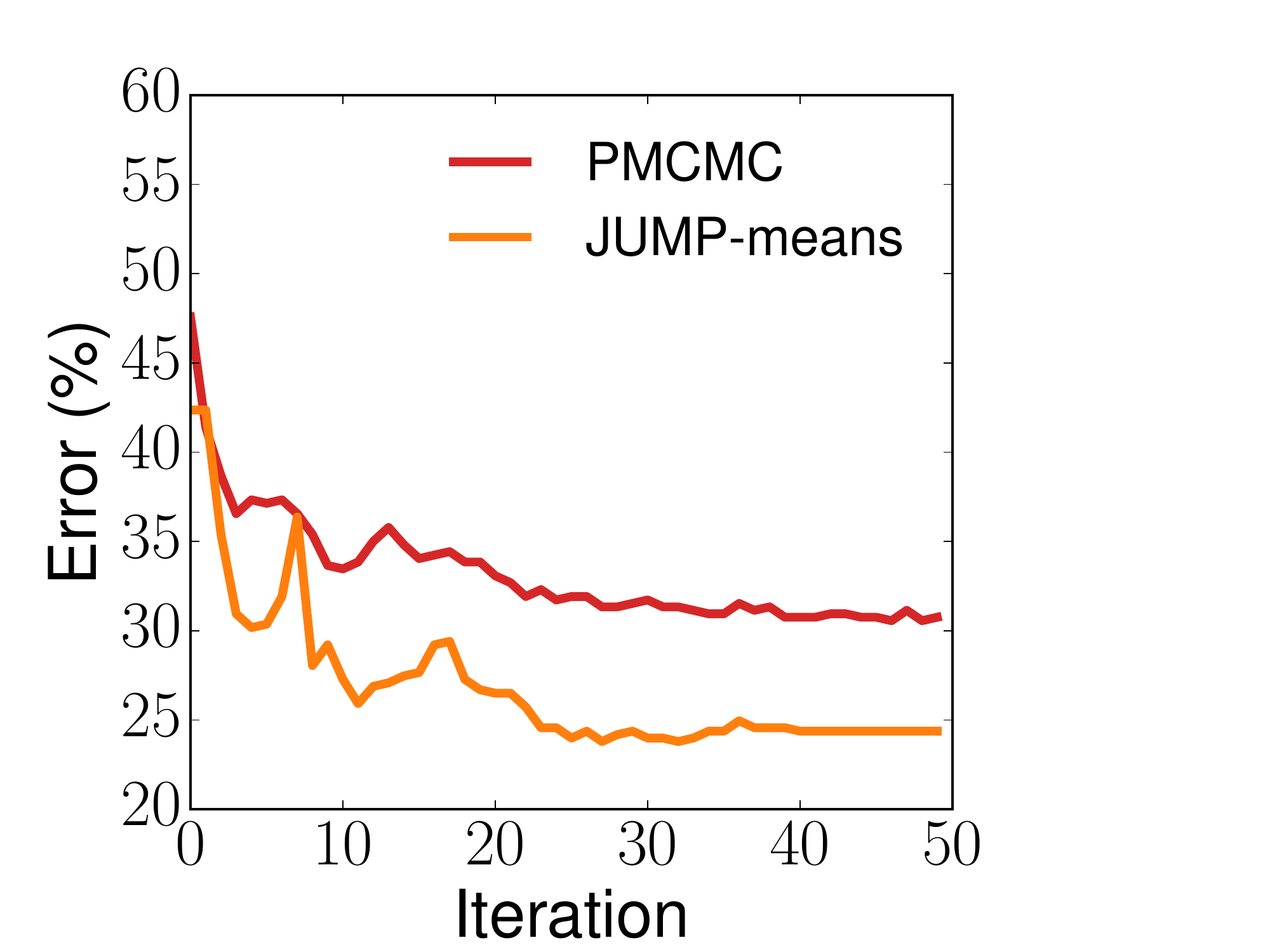}%
}
\vspace{-4mm}
\caption{\footnotesize{Mean error vs iterations for 
\textbf{(a)} Synthetic 1; \textbf{(b)} Synthetic 2; \textbf{(c)} MS; and \textbf{(d)} MIMIC datasets.
In each case the JUMP-means algorithms have better or comparable performance to 
other standard methods of inference in MJPs.
Mean error vs CPU runtime plots can be found in the Supplementary Material.}}
\label{fig:time}
\vspace{-4mm}
\end{figure*}

\brmk
If instead of multinomial observations we have Gaussian observations,
the parameter $\boldsymbol\rho_{s}$ is replaced with the mean parameter $\mu_{s}$. 
In this case, we update the mean for each state using the data points 
assigned to the state, similar to the procedure for $k$-means 
clustering (see, e.g., \citet{jiang2012small,Roychowdhury:2013}).
\ermk

\section{Experiments}
\label{sec:experiments}
In this section we provide a quantitative analysis of the JUMP-means algorithm 
and compare its performance on synthetic and real datasets with 
standard inference methods for MJPs. 
For evaluation, we consider multiple sequences of discretely observed 
data and randomly hold out a subset of the data. 
We report reconstruction error for performance comparison. 

\subsection{Parametric Models}
For the parametric models, we compare JUMP-means to maximum 
likelihood estimation of the MJP parameters learned by 
EM~\citep{asger2005statistical}, the MCMC method proposed 
in~\citet{Rao:2013} and a simple baseline where we ignore the 
sequential structure of the data. 
We run three sets of experiments (2 synthetic, 1 real) for our evaluation.

\textbf{Synthetic 1: Directly Observed States} For evaluating the model 
on a directly observed process, we generate 100 different datasets randomly
from various MJPs with 10 states. 
To generate each dataset, we first generate the rows of the transition 
probability matrix and transition rates independently from $\distDir(1)$ 
and $\distGam(1, 1)$, respectively. 
Next, given the rates and transition probabilities for each dataset, 
we sample 500 sequences of length 20. 
We hold out 30\% of the observations at random for testing reconstruction error.

We run JUMP-means by initializing the algorithm with a uniform transition 
matrix $P$ and set the rate vector $\boldsymbol\lambda$ to be all ones. 
We run 300 iterations of the algorithm described in Section \ref{sec:parametric_alg}; 
each iteration is one scan through all the sequences. 
We set the hyperparameters $\xi,\xi_{\lambda}$, and $\mu_{\lambda}$
equal to 1, 1, and .5, respectively.
For MCMC, we initialize the jump points using the 
time points of the observations. 
We place independent $\distDir(1)$ priors on $P$ and independent 
$\distGam(1, 1)$ priors on $\boldsymbol\lambda$. 
We initialize EM with a uniform $P$ and an all-ones 
$\boldsymbol\lambda$. 
We run both MCMC and EM for 300 iterations,
then reconstruct observations using the Bayes estimator approximated 
from the 300 posterior samples.  
For our baseline we use the most common observation in the dataset as an 
estimate of the missing observations. 
 
Table \ref{tbl:experiments} gives the mean reconstruction error across sequences
for the various methods. 
Note that JUMP-means performs better than MCMC, and is almost on par with EM. 
Fig. \ref{fig:time}(a) shows the average error across all the datasets for 
each method versus number of iterations.  
In terms of CPU time, each iteration of JUMP-means (Java), EM (Java), 
and MCMC (Python) takes 0.3, 1.61 and 42 seconds, respectively. 
We also ran experiments with the beam search variant described in 
Section \ref{sec:parametric_alg}; however, we did not obtain any significant 
improvement in results.  

\textbf{Synthetic 2: Hidden States} For the hidden state case, we generate 100 different datasets for MJPs with 5 hidden and 5 observed states, with varying parameters as above. In each dataset there are 500 sequences of length 20. In addition to parameters in the directly observed case, we generate observation likelihood terms for each state from $\distDir(1)$. 

We initialize the transition probabilities and the rate vectors for JUMP-means, MCMC and EM in a fashion similar to the directly observed case.  For the observation likelihood $\boldsymbol\rho$, we use $\distDir(1)$ as a prior for MCMC, uniform distributions for EM initialization and a uniform probability matrix with a small amount of random noise for JUMP-means initialization. We set $\xi,\xi_{\lambda},\mu_{\lambda}$ as before and $\xi$ to 1.

We run each algorithm for 300 iterations. For JUMP-means, we use the hidden state MJP algorithm described in Section \ref{sec:parametric_alg}.  Table \ref{tbl:experiments} and Fig. \ref{fig:time}(b) again demonstrate that JUMP-means outperforms MCMC by a large margin and performs comparably to EM. The poor performance of MCMC is due to slow mixing over the parameters and state  trajectories. The slow mixing is a result of the coupling between the latent states and the observations, which is induced by the observation likelihood. 

\textbf{Disease Progression in Multiple Sclerosis (MS)}  Estimating disease progression and change points in patients with Multiple Sclerosis (MS) is an active research area (see, e.g.,~\citet{mandel2010estimating}). We can cast the progression of the disease in a single patient as an MJP, with different states representing the various stages of the disease. Obtaining the most-likely trajectory for this MJP can aid in understanding the disease progression and  enable better care. 

For our experiments, we use a real-world dataset collected from a phase III clinical trial of a drug for MS. This dataset tracks the progression of the disease for 72 different patients over three years. We randomly hold out 50$\%$ of the observations and evaluate on the observation reconstruction task. The observations are values of a disability measure known as EDSS, recorded at different time points. Initialization and hyperparameters are the same as Synthetic 1.

\begin{figure}[tb]
\vspace{-2mm}
\begin{center}
\includegraphics[width = \columnwidth]{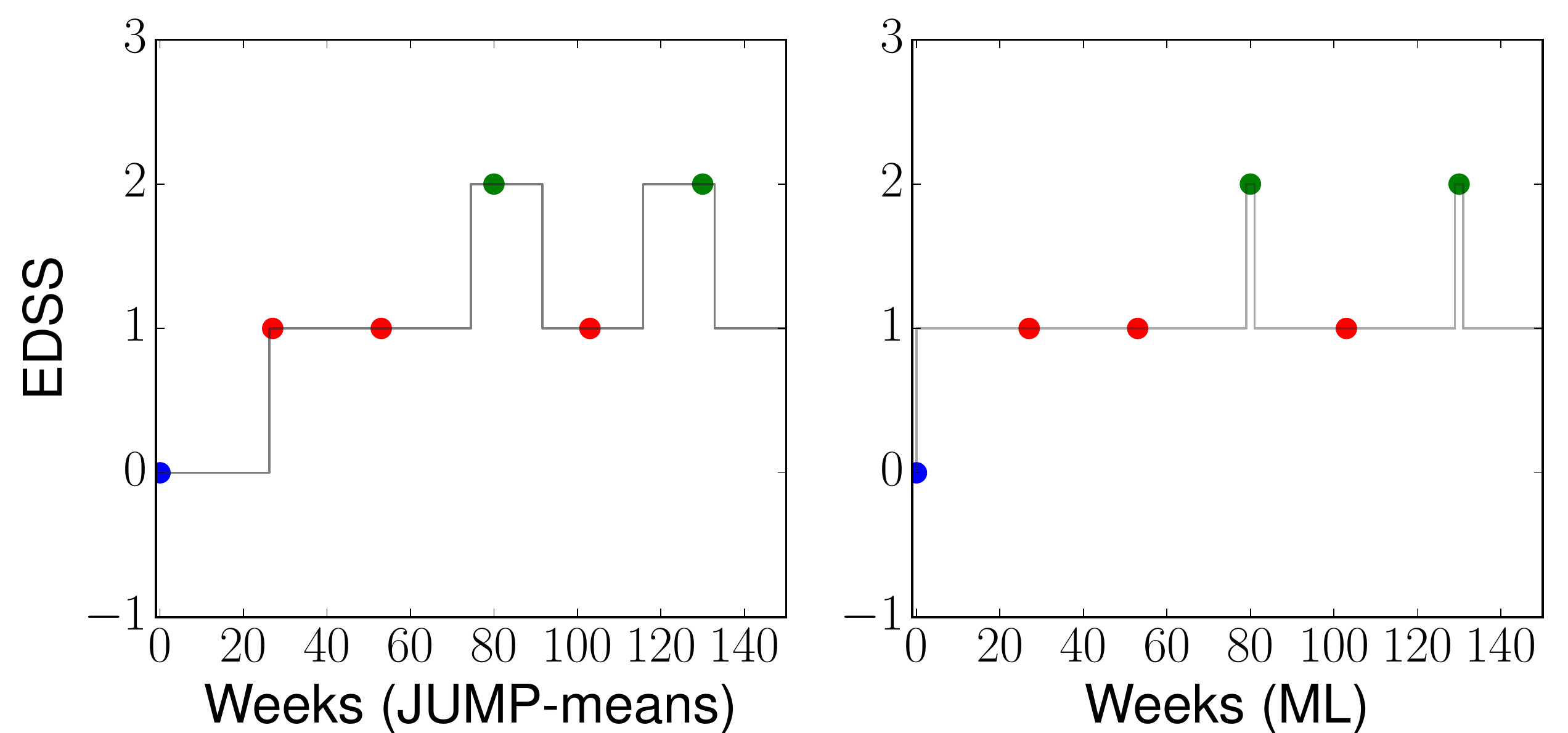} \\ 
\includegraphics[width = \columnwidth]{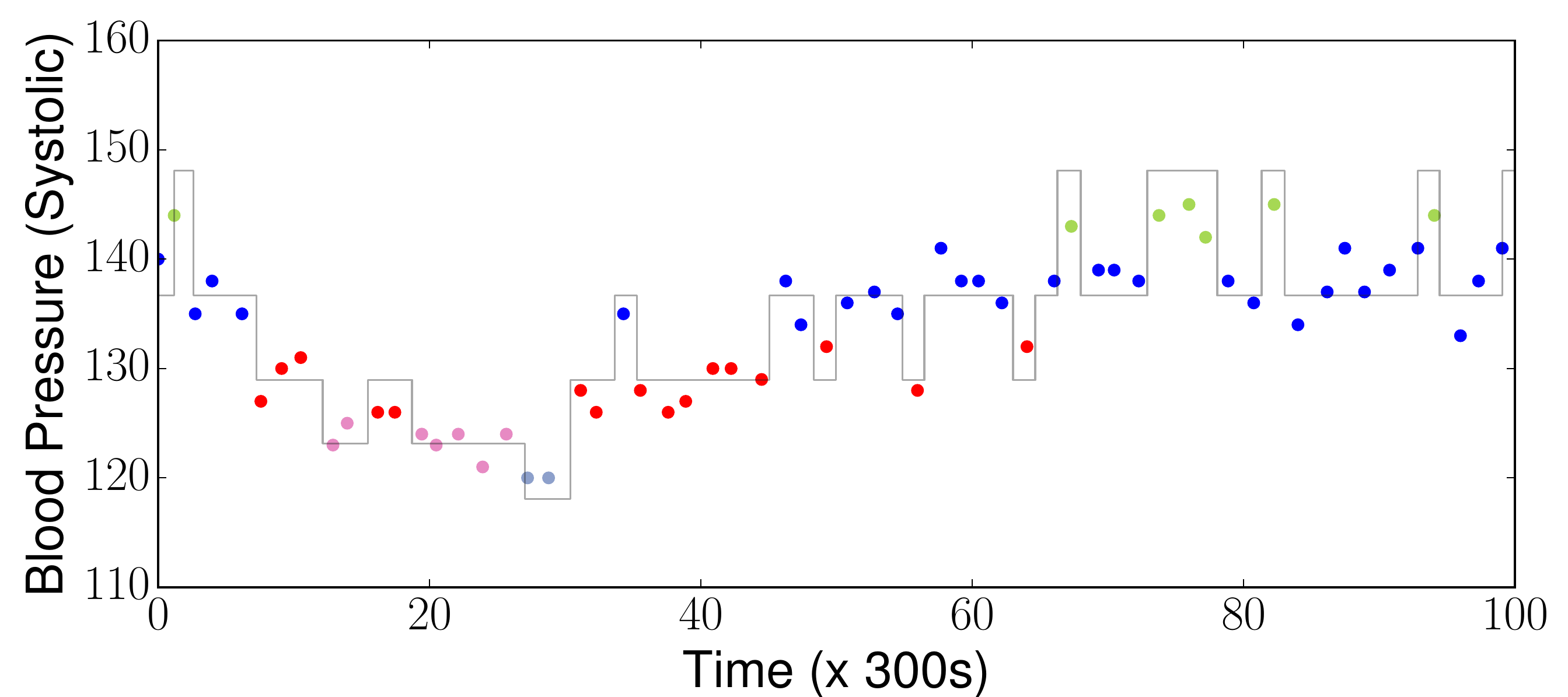} 
\vspace{-8mm}
\caption{\textbf{Top:} Latent trajectories inferred by JUMP-means and ML estimate 
for a patient in the MS dataset. 
\textbf{Bottom:} Latent trajectory inferred by JUMP-means for a patient in MIMIC dataset.}
\label{fig:npb}
\end{center}
\vspace{-8mm}
\end{figure}

Table \ref{tbl:experiments} shows that JUMP-means significantly outperforms MCMC, achieving almost a 50\% relative reduction in reconstruction error. JUMP-means again achieves comparable results with EM.
Fig. \ref{fig:npb} (top panel) provides an example of the latent trajectories from JUMP-means  and maximum likelihood estimate for a single patient. The MLE trajectory includes two infinitesimal dwell times, which do not reflect realistic behavior of the system (since we do not expect a patient to be in a disease state for an infinitesimal amount of time). On the other hand, the trajectory produced by JUMP-means takes into account the dwell times of the various stages of the disease and provides a more reasonable picture of its progression.

\subsection{Nonparametric Model}

\textbf{Vital Signs Monitoring (MIMIC)}
We now consider a version of the problem of understanding 
physiological signals discussed in the introduction. 
We use data from the MIMIC database~\citep{PhysioNet,moody1996mimic}, 
which contains recordings of several vital parameters of ICU patients. 
Specifically, we consider blood pressure readings 
of 69 ICU patients collected over a 24-hour period and sub-sample 
observation sequences of length 32 for each patient, keeping the 
start and end times fixed.\footnote{We use a small dataset for testing since 
PMCMC cannot easily scale to larger datasets.}
For testing, we randomly hold out $\sim$25\% of the observations.

To initialize JUMP-means, we choose uniform matrices for 
$\boldsymbol\rho, \bar\pi_0$ and $\bar\pi_1$ and set $M$ = 1. 
The hyperparameters $\gamma$ and $\xi_1$ are set to 5, 
while $\zeta, \xi$, and $\xi_2$ are set to 0.005.
Using a Gaussian likelihood model for the observations, 
we run our model for 50 iterations.
We compare with particle MCMC (PMCMC)~\citep{Andrieu:2010} and EM.
PMCMC is a state-of-the-art inference method for iMJPs \citep{saeedi2011priors}, 
which we run for 300 iterations with 100 particles. 
For PMCMC, we first categorize the readings into the
standard four categories for blood pressure provided by 
NIH\footnote{\url{http://www.nhlbi.nih.gov/health/health-topics/topics/hbp}}. 
We run EM with a number of hidden states 
from 1 to 12 and report the best performance among all the results. 
For initializing the EM, we use the same setting as the Synthetic 2 case. 

\begin{figure}[bt]
\vspace{-2mm}
\begin{center}
\hspace{-4mm}
\includegraphics[width=0.48 \columnwidth ]{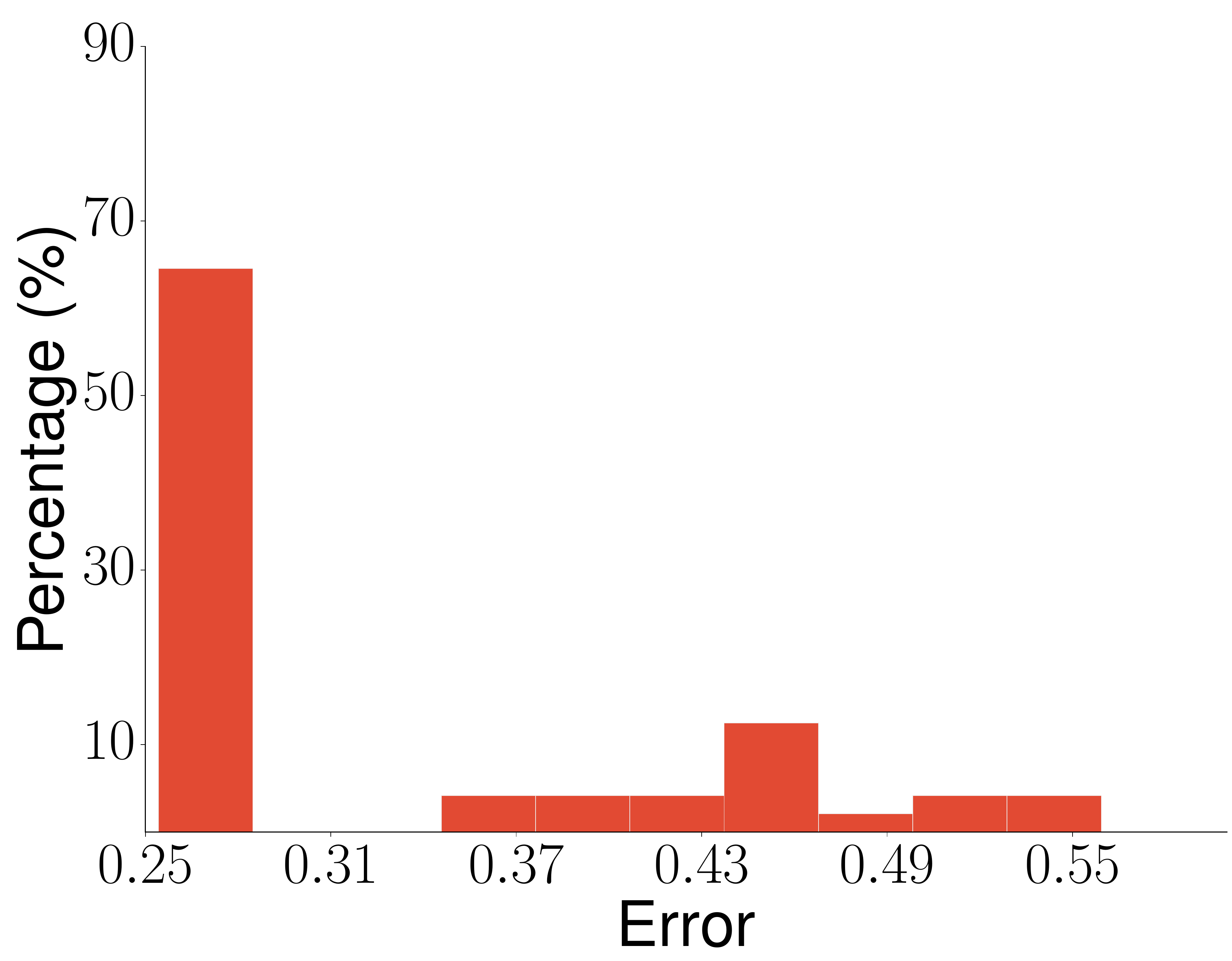}  
\includegraphics[width= 0.45 \columnwidth]{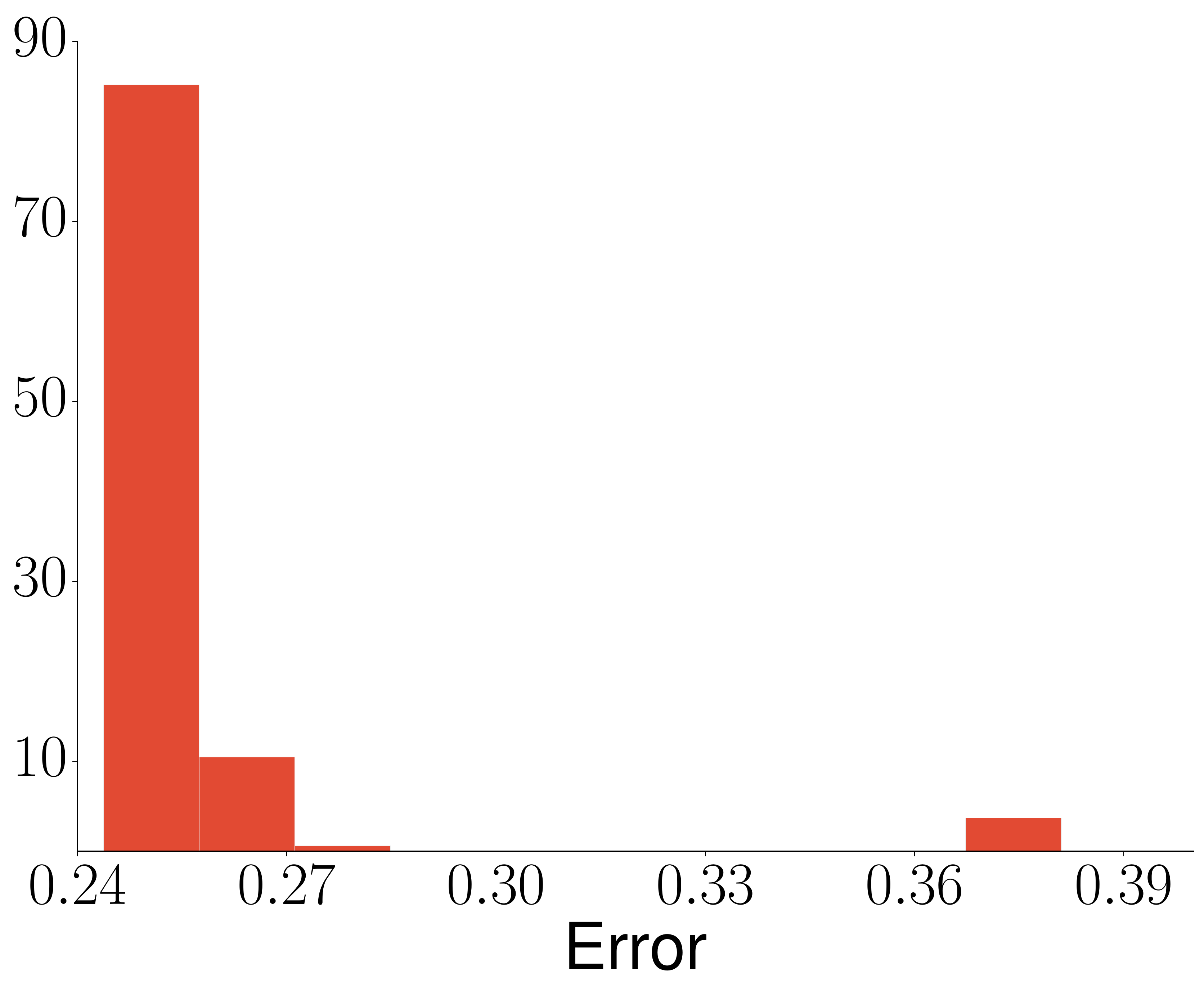} 
\vspace{-2mm}
\caption{Histograms of error reconstruction for runs with different hyperparameter settings for (a) MS (P-DO, 48 settings), and (b) MIMIC (NPB-H, 1125 settings) datasets.}
\label{fig:hyperparams}
\end{center}
\vspace{-6mm}
\end{figure}

For evaluation, we consider the time point of a test observation and 
categorize the mean of the latent state at this time point  
(using the same categories obtained above) to compare against the actual category. 
Table \ref{tbl:experiments} shows that JUMP-means significantly outperforms 
PMCMC and obtains a 21\% relative reduction in average error rate. 
Fig. \ref{fig:time}(c) plots the error against iterations of both algorithms. 
In terms of CPU time, each iteration of JUMP-means (Java) and PMCMC (Java)
takes 0.17 and 1.95 seconds, respectively. 
Compared to EM's error rate of 25.7\%, JUMP-means reaches a 
rate of 24.3\% without the need 
to separately train for different number of states. 
The second-best result for the EM had an error of 45\%, 
which shows the importance of model selection when using EM.

Fig. \ref{fig:npb} (bottom) provides an example of the latent trajectory inferred by JUMP-means. The observations are uniquely colored by the latent state they are assigned. We note that the model captures different levels of blood pressure readings and provides a non-degenerate latent trajectory.

\textbf{Hyperparameters} A well-known problem when applying SVA methods is 
that there are a number of hyperparameters to tune. 
In our objective functions, some of these hyperparameters 
($\gamma$, $\mu_\lambda$, and $\xi_\lambda$) have natural interpretations 
so prior knowledge and common sense can be used to set them, but others do not. 
Fig. \ref{fig:hyperparams} shows histograms over the errors we obtain for runs of JUMP-means on 
the MS and MIMIC datasets with different settings. 
We can see that a significant fraction of the runs converge to the minimum error, 
while some settings --- in particular when the hyperparameters were of 
different orders of magnitude --- led to larger errors. 
Hence, the sensitivity study indicates the robustness of JUMP-means 
to the choice of hyperparameters.

\begin{figure}[bt]
\vspace{-3mm}
\begin{center}
\includegraphics[width=0.9\columnwidth ]{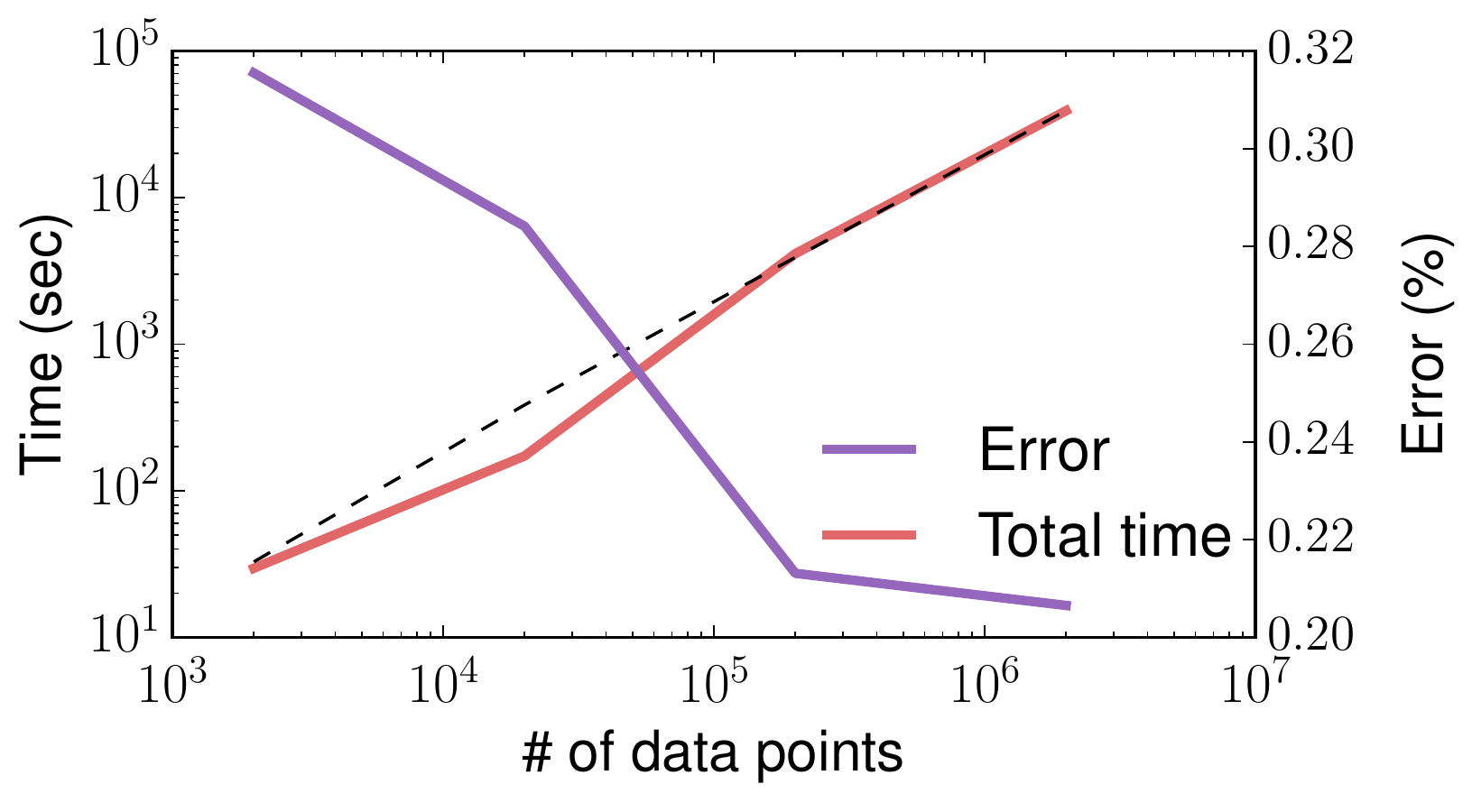} 
\vspace{-3mm}
\caption{Runtime and error of nonparametric JUMP-means algorithm with 
increasing synthetic data size.
The runtime scales linearly with data size (dashed black line).  
}
\label{fig:computation}
\end{center}
\vspace{-8mm}
\end{figure}

\textbf{Scaling} Fig. \ref{fig:computation} shows the total runtime 
 and reconstruction error of the non-parametric JUMP-means
algorithm on increasingly large amounts of synthetic data. 
The algorithm is able to handle up to a million data points 
with the runtime scaling linearly with data size. 
Furthermore, the error rate decreases significantly as the amount of data
increases. See the Supplementary Material for further details.

\vspace{-2.6mm}

\section{Conclusion}

We have presented JUMP-means, a new approach to inference 
in MJPs using small-variance asymptotics. 
We derived novel objective functions for parametric 
and Bayesian nonparametric models and proposed efficient 
algorithms to optimize them. 
Our experiments demonstrate that JUMP-means
can be used to obtain high-quality non-degenerate estimates 
of the latent trajectories.
JUMP-means offers 
attractive speed-accuracy tradeoffs for both
parametric and nonparametric problems, and 
achieved state-of-the-art reconstruction 
accuracy on nonparametric problems.

\subsubsection*{Acknowledgments}
Thanks to Monir Hajiaghayi, Matthew Johnson, and Tejas Kulkarni for helpful comments and discussions.
JHH was supported by the U.S. Government under FA9550-11-C-0028 and awarded by the DoD, Air Force
Office of Scientific Research, National Defense Science and Engineering Graduate
(NDSEG) Fellowship, 32 CFR 168a.

\bibliographystyle{icml2015}
\bibliography{sva-mjp,sva-mjp-jhh}

\newpage
\onecolumn

\boldshortcuts

\allowdisplaybreaks[3]

\title{Supplementary Material for \\ \emph{JUMP-Means: Small-Variance Asymptotics for \\ Markov Jump Processes}}


\date{}
\maketitle

\appendix 

\numberwithin{equation}{section}

\section{Parametric MJPs for SVA}

To obtain the SVA objective from the parametric MJP 
model, we begin by scaling the exponential distribution 
$f(t;\lambda)=\lambda\exp(-\lambda t)$, which is an
exponential family distribution with natural parameter $\eta = -\lambda$,
log-partition function $\psi(\eta) = - \ln(-\eta)$,
and base measure $\nu(\dee t) = 1$~\citep{banerjee2005clustering}. 
To scale the distribution, introduce the
new natural parameter $\tilde\eta = \beta\eta$ and
log-partition function $\tilde\psi(\tilde\eta) = \beta\psi(\tilde\eta/\beta)$.
The new base measure $\tilde\nu(\dee t)$ is uniquely defined by 
the integral equation~\citep[see][Theorem 5]{banerjee2005clustering}
\(
\int \exp(\tilde\eta t)\tilde\nu(\dee t)
= \exp(\tilde\psi(\tilde\eta)) 
= \exp(-\beta \ln(\tilde\eta/\beta) )
= \frac{\beta^{\beta}}{{\tilde\eta}^{\beta}}.
\)
Choosing $\tilde\nu(\dee t) = \frac{t^{\beta - 1}\beta^{\beta}}{\Gamma(\beta)}\dee t$
satisfies the condition, so we have
\(
f(t;\lambda,\beta) 
= \frac{(\beta\lambda)^{\beta}}{\Gamma(\beta)}t^{\beta - 1}e^{-\beta\lambda t}
&= \exp(-\beta \lambda t + (\beta - 1) \ln t + \beta \ln \lambda \beta - \ln \Gamma(\beta)) \\
&= \exp\left\{ -\beta\left( 
      \lambda t - \ln t  - \ln \lambda 
    - \frac{\beta \ln \beta - \ln \Gamma(\beta)}{\beta} 
    + \frac{\ln t}{\beta}  \right)\right\}.
\)
It can now be seen that $f(t;\lambda,\beta)$ is the density of a
gamma distribution with shape parameter $\beta$ and rate 
parameter $\beta\lambda$.
Hence, the mean of the scaled distribution is $\frac{1}{\lambda}$ 
and its variance is $\frac{1}{\lambda\beta}$. 
Letting $F(t; \lambda, \beta)$ denote the CDF corresponding to $f(t; \lambda,\beta)$, 
we have $1-F(t; \lambda,\beta) = \frac{\Gamma(\beta, \beta \lambda t)}{\Gamma(\beta)}$,
where $\Gamma(\cdot,\cdot)$ is the upper incomplete gamma function. 

For the state at the $k$-th jump we use a 1-of-$M$ representation; 
that is, $s_k$ is an $M$-dimensional binary random variable which 
satisfies $s_{km} \in \{0, 1\}$ and $\sum_{m=1}^{M} s_{km} = 1$. 
Hence, we have:
\begin{align}
p(s_{k} | s_{k-1,j}=1) = \prod^M_{m=1} p_{jm}^{s_{km}}.
\end{align}
Given the Bregman divergence for a multinomial distribution, 
$d_{\phi}(s_{k}, \bp_j) = \kl{s_{k}}{\bp_j}$ where $\bp_j \defined (p_{j1},\dots,p_{jM})$, 
this can be written in terms of exponential family notation in 
the following form~\citep{banerjee2005clustering}:
\begin{align}
p(s_{k} | s_{{k-1}, j} = 1) = b_{\phi}(s_{k})\exp(-d_{\phi}(s_{k}, \bp_j))
\end{align}
where $b_{\phi}(s_k) = 1$. 
For a scaled multinomial distribution we have 
$b_{\hat{\beta} \phi}(s_{k})\exp(-\hat{\beta} d_{\phi}(s_{k}, \bp_j))$,
where $\hat{\beta} = \xi \beta$ is the scaling parameter for the 
multinomial distribution. 
Writing the trajectory probility with the scaled exponential 
families yields:
\[
\begin{split}
p(\mcU|s_0, s_{K}, P, \boldsymbol\lambda)
&\propto \exp \left\{
  - \beta \left(\frac{\ln \Gamma(\beta) 
  - \ln\Gamma(\beta, \beta \lambda_{s_{K}}t_{\cdot})}{\beta} 
  + \xi \sum^{K-1}_{k=0} \kl{s_{k+1}}{\bp_{s_{k}}} \right. \right. \\
&\phantom{\propto \exp\{} \left. \left.  
   + \sum^{K-1}_{k=0}\left(\lambda_{s_{k}} t_k - \ln \lambda_{s_{k}}t_k - \frac{\beta \ln \beta - \ln \Gamma(\beta)}{\beta} + \frac{\ln t_k}{\beta} \right)\right)\right\},
\end{split}
\]
Since $\beta \to \infty$, we can apply the asymptotic 
expansions for $\Gamma(\cdot)$ and $\Gamma(\cdot,\cdot)$.
In particular, applying Stirling's formula and 
the facts in~\citep{NIST:DLMF} we have: 
\(
\frac{\beta \ln \beta - \ln \Gamma(\beta) }{\beta} 
&= \frac{\beta\ln \beta -\beta\ln\beta + \beta + o(\beta)}{\beta} \to 1 \\
\frac{\ln \Gamma(\beta) - \ln\Gamma(\beta, \beta \lambda t )}{\beta} 
&= \left\{
                \begin{array}{ll}
                  \frac{- \beta - o(\beta)  - \beta \ln \lambda t + \beta \lambda t}{\beta} \to \lambda t - \ln \lambda t - 1 &\text{\hspace{0.05in} if  $t \geq \frac{1}{\lambda}$}\\
                  \frac{\beta\ln \beta - \beta -\beta\ln\beta + \beta + o(\beta)}{\beta} \to 0 &\text{\hspace{0.05in} if $t < \frac{1}{\lambda}$} 
                \end{array}
              \right.
\)

We also place a $\distGam(\alpha_{\lambda}, \alpha_{\lambda}\mu_{\lambda})$ 
prior on each $\lambda_{i}$.
With $\alpha_{\lambda} = \xi_{\lambda}\beta$, we 
obtain
\(
\ln p(\lambda_{s} \given \alpha_{\lambda}, \alpha_{\lambda}\mu_{\lambda})
&= \alpha_{\lambda}\ln(\alpha_{\lambda}\mu_{\lambda}) 
   + (\alpha_{\lambda}-1)\ln \lambda_{s} 
   - \ln \Gamma(\alpha_{\lambda}) 
   - \alpha_{\lambda}\mu_{\lambda}\lambda_{s} \\
&= \xi_{\lambda}\beta\ln \lambda_{s} - \xi_{\lambda}\mu_{\lambda}\beta\lambda_{s} + \xi_{\lambda}\beta + o(\beta)  \\
&= - \beta(\xi_{\lambda}\mu_{\lambda}\lambda_{s} - \xi_{\lambda}\ln \lambda_{s} - 1) + o(\beta). 
\)
Hence, when $\beta \to \infty$, obtain
\[
\begin{split}
\min_{\mcU, \boldsymbol\lambda, P} \bigg\{
&   \xi \sum^{K-1}_{k=0}\kl{s_{k+1}}{\bp_{s_{k}}} 
  + \sum^{K-1}_{k=0}\left(\lambda_{s_k} t_k - \ln \lambda_{s_k} t_k - 1\right) \\
& + \ind[\lambda_{s_{K}}t_{\cdot} \geq 1](\lambda_{s_{K}}t_{\cdot} - \ln \lambda_{s_{K}}t_{\cdot} - 1)
  + \xi_{\lambda}\sum_{s=1}^{M}(\mu_{\lambda}\lambda_{s} - \ln \lambda_{s} - 1) \bigg\}
\end{split} \label{eq:mjp-objective}
\]

\section{Bayesian Nonparametric MJPs for SVA}

First we recall that the Moran gamma process is a distribution over measures.
If $\mu \dist \distGamP(H, \gamma)$ is a random measure distributed 
according to a Moran gamma process with base measure $H$ on the 
probability space $(\Omega, \mcF)$ and rate parameter $\gamma$, 
then for all measurable partitions of $\Omega$, $(A_{1},\dots,A_{\ell})$, 
$\mu$ satisfies
\[
(\mu(A_{1}),\dots,\mu(A_{\ell}))
&\dist \distGam(H(A_{1}), \gamma) \times \dots \times \distGam(H(A_{\ell}), \gamma).
\]
The hierarchical gamma-gamma process (\HGGP) is defined to be:
\[
\mu_{0} &\dist \distGamP(\alpha_{0}H_{0}, \gamma_{0}) \\
\mu_{i} &\given \mu_{0} \distiid \distGamP(\beta\mu_{0}, \gamma) & i = 1,2,\dots \label{eq:hggp-mu-i} \\
s_{k} &\given \{\mu_{i}\}_{i=0}^{\infty}, \mcU_{k-1} \dist \bar\mu_{s_{k-1}}  \\
t_{k} &\given  \{\mu_{i}\}_{i=0}^{\infty}, \mcU_{k-1} \dist \distGam(\beta, \|\mu_{s_{k-1}}\|). \label{eq:hggp-time-dist}
\]
Consider the gamma-gamma process (\GGP), defined by 
\eqref{eq:hggp-mu-i}-\eqref{eq:hggp-time-dist} 
(with $\mu_{0}$ treated as an arbitrary fixed measure).
We now show that the \GGP retains the key properties of the $\Gamma$EP:
conjugacy and exchangeability. 
Let $T_{i} \defined \sum_{j=1}^{k} \ind[s_{j-1} = i]t_{j}$ 
and $F_{i} \defined \sum_{j=1}^{k} \ind[s_{j-1} = i]\delta_{s_{j}}$
be the sufficient statistics of the observations. 
\bnprop \label{prop:ggp-conjugacy}
The \GGP is a conjugate family: 
$\mu_{i} \given \mcU_{k} \dist \distGamP(\beta\mu_{i}', \gamma_{i}')$,
where $\mu_{i}' = \mu_{0} + F_{i}$ and $\gamma_{i}' = \gamma + T_{i}$. 
\enprop
\bprf[Proof sketch] 
The proof is analogous to that for Proposition 2 in \citep{saeedi2011priors}. 
The key additional insight is that $X \dist \distGam(\beta a,b)$ and 
$Y \given X \dist \distGam(\beta, X)$ are conjugate: 
$X \given Y \dist \distGam(\beta(a + 1), b + Y)$. 
\eprf
In order to give the joint distribution of the times 
$\mcT \defined \mcT_{K} \defined (t_{1},\dots, t_{K})$,
we first derive the predictive distribution for the \GGP, 
$(s_{k+1}, t_{k+1}) \given \mcU_{k}$. 
We make use of the following family of densities. 
\bnumdefn[Shaped Translated Pareto]
Let $\beta > 0,\alpha > 0, \gamma >0$. 
A random variable $S$ is \emph{shaped translated Pareto},
denoted $S \dist \distSTP(\beta, \alpha, \gamma)$, if it has
density
\(
f(t) 
= \frac{\gamma^{\alpha\beta}}{B(\beta,\alpha\beta)}\frac{t^{\beta - 1}}{(t + \gamma)^{(1 + \alpha)\beta}},
\)
where $B(a,b) = \frac{\Gamma(a)\Gamma(b)}{\Gamma(a + b)}$ is
the beta function. 
\enumdefn
\bnprop 
The predictive distribution of the \GGP is 
\[
(s_{k+1}, t_{k+1}) \given \mcU_{k} 
\dist \bar\mu_{s_{k}}' \times \distSTP(\beta, \|\bar\mu_{s_{k}}'\|, \gamma_{s_{k}}'). 
\]
\enprop
\bprf
By Proposition \ref{prop:ggp-conjugacy}, it suffices to show 
that if $\mu \dist \distGamP(\beta\mu_{0}, \gamma)$, 
$s \given \mu \dist \bar\mu$, and 
$t \given \mu \dist \distGam(\beta, \|\mu\|)$, 
then $(s, t) \dist \bar\mu \times \distSTP(\beta, \kappa_{0}, \gamma)$,
where $\kappa_{0} \defined \|\mu_{0}\|$. 
Letting $x = \|\mu\|$, the distribution of $t$ is 
\(
p(t) 
&= \int_{0}^{\infty} p(t \given x)p(x) \dee x
= \int_{0}^{\infty} \frac{x^{\beta}t^{\beta-1}e^{-xt}}{\Gamma(\beta)}
  \frac{\gamma^{\beta\kappa_0}x^{\beta\kappa_0-1}e^{-\gamma x}}{\Gamma(\beta\kappa_0)} \dee x \\
&= \frac{\gamma^{\beta\kappa_0}t^{\beta-1}}{\Gamma(\beta)\Gamma(\beta\kappa_0)} 
   \int_{0}^{\infty} x^{\beta(1 + \kappa_0) - 1}e^{-(\gamma + t)x} \dee x
= \frac{\gamma^{\beta\kappa_0}t^{\beta-1}}{\Gamma(\beta)\Gamma(\beta\kappa_0)} 
  \frac{\Gamma(\beta(1 + \kappa_0))}{(\gamma + t)^{\beta(1 + \kappa_0)}}.
\)
\eprf
We can now show that the process is exchangeable by
exhibiting the joint distribution of waiting times:
\bnprop
Let $\bt^{*}_{m} = (t^{*}_{m1},\dots,t^{*}_{mK_{m}})$ be the 
waiting times following state $m$. 
Then $\bt^{*}_{m}$ is an exchangeable sequence with joint
distribution 
\[
p(\bt^{*}_{m}) 
= \frac{\Gamma(\beta(\kappa_0 + K_{m}))}{\Gamma(\beta)^{K_{m}}}
  \frac{(\prod_{j=1}^{K_{m}}t^{*}_{mj})^{\beta - 1}}{(\gamma + \sum_{j=1}^{K_{m}} \tau_{mj})^{\beta(\kappa_0 + K_{m})}}
\]
\enprop
\bprf[Proof sketch] 
Take the product of the predictive distributions of 
$\tau_{m1},\dots,\tau_{mK_{m}}$. 
\eprf

The measures $\{\mu_{i}\}_{i=0}^{\infty}$ and $H_{0}$ 
can be integrated out of the \HGGP generative model
in a manner analogous to the way in the the Chinese
restaurant franchise in obtained from the 
hierarchical Dirichlet process \citep{Teh:2006b}.
However the mass of the measure $\mu_{0}$ cannot 
be integrated out.
We omit details as they are essentially identical to 
those in case of the H$\Gamma$EP \citep{saeedi2011priors}.

First, we consider the case of integrating out 
$\{\mu_{i}\}_{i \ge 0}$. 
Let $M$ denote the number of used states, $K_{m}$ the number
of transitions out of state $m$, and $r_{m}$ the number of 
states that can be reached from state $m$ in one step.
The contribution to the
likelihood from the \HGGP prior is
\(
p(\mcU, \kappa_{0} \given \beta,  \gamma_{0}, \gamma, \alpha_{0})
&= p(\kappa_{0} \given \alpha_{0}, \gamma_{0}) 
   p(\mcS \given  \beta, \alpha_{0}, \kappa_{0}) 
   p(\mcT \given   \beta, \gamma, \kappa_{0})\\
&\propto 
   \kappa_{0}^{\alpha_{0}-1}e^{-\gamma_{0}\kappa_{0}}
   \alpha_{0}^{M-1}\frac{\Gamma(\alpha_{0} + 1)}{\Gamma(\alpha_{0} + r_{\cdot})} 
   \prod_{m=1}^{M}(\beta\kappa_0)^{r_{m} - 1} \frac{\Gamma(\beta\kappa_{0} + 1)}{\Gamma(\beta\kappa_0 + K_{m})} \\
&\phantom{\propto}
  \times \prod_{m=1}^{M} \frac{\Gamma(\beta(\kappa_0 + K_{m}))}{\Gamma(\beta)^{K_{m}}}
  \frac{(\prod_{j=1}^{K_{m}}t^{*}_{mj})^{\beta - 1}}{(\gamma + \sum_{j=1}^{K_{m}} t^{*}_{mj})^{\beta(\kappa_0 + K_{m})}},
\)
where $r_{\cdot} \defined \sum_{m} r_{m}$. 
Taking the logarithm, using asymptotic expansions for the 
Gamma terms, and ignoring $o(\beta)$ terms yields
\(
&(\alpha_{0} - 1)\ln \kappa_{0} - \gamma_{0}\kappa_{0} + (M - 1)\ln\alpha_{0} + 
  \sum_{m=1}^{M} \left\{(r_{m} - 1)\ln \kappa_{0} + \beta(\kappa_{0} + K_{m})\ln[\beta(\kappa_{0} + K_{m})] \right\} \\
&\sum_{m=1}^{M} \left\{ - \beta(\kappa_{0} + K_{m}) - K_{m}[\beta \ln \beta - \beta] 
 + \beta\textstyle\sum_{j=1}^{K_{m}}\ln t^{*}_{mj} 
 -  \beta(\kappa_{0} + K_{m})\ln\left(\gamma +  t^{*}_{m\cdot}\right) \right\},
\)
where $t^{*}_{m\cdot} \defined \sum_{j=1}^{K_{m}}t^{*}_{mj}$. 
In order to retain the effects of the hyperparameters in the 
asymptotics, set $\alpha_{0} = \exp(-\xi_{1}\beta)$ and
$\gamma_{0} = \exp(\xi_{2}\beta)$. 
Thus, $\kappa_{0} \to 0$ as $\beta \to \infty$. 
We require that 
$\limsup_{\beta \to \infty}\kappa_{0}\gamma_{0} < \infty$,
so without loss of generality we can choose 
$\kappa_{0} = \gamma_{0}^{-1} = \exp(-\xi_{2}\beta)$
to obtain
\(
-\beta\left(\xi_{1}(M - 1) + \sum_{m=1}^{M} \left\{\xi_{2}(r_{m} - 1) - \textstyle\sum_{j=1}^{K_{m}}\ln t^{*}_{mj} 
 + K_{m}\ln \left([\gamma + t^{*}_{m\cdot}]/K_{m}\right) \right\}\right).
\)
Thus, the objective function to minimize is
\[
&\phantom{+}~\zeta\sum_{\ell=1}^{L} \kl{x_{\ell}}{\boldsymbol\rho_{s_{\tau_{\ell}}}} 
+ \xi_{1}M 
+ \sum_{m=1}^{M} \left\{\xi_{2}(r_{m} - 1) - \textstyle\sum_{j=1}^{K_{m}}\ln t^{*}_{mj} 
- K_{m}\ln \left([\gamma + t^{*}_{m\cdot}]/K_{m}\right) \right\}.
\]

Alternatively, the small variance asymptotics can be 
derived in the case where $\{\mu_{i}\}_{i \ge 0}$ is not
integrated out.
To do so, we first rewrite the \HGGP generative model in 
an equivalent form, with $H_{0}$ integrated out:
\[
\pi_{0} &\dist \distNamed{GEM}(\alpha_{0}) \\
\kappa_{0} &\dist \distGam(\alpha_{0}, \gamma_{0}) \\
\pi_{i} &\given \pi_{0} \distiid \distDP(\beta\kappa_{0}\pi_{0}), & i = 1,2,\dots \\
\kappa_{i} &\given \pi_{0} \distiid \distGam(\beta, \gamma), & i = 1,2,\dots \\
s_{k} &\given \{\pi_{i}\}_{i=1}^{\infty}, \mcU_{k-1} \dist \pi_{s_{k-1}} \label{eq:hggp-state-dist} \\
t_{k} &\given  \{\kappa_{i}\}_{i=1}^{\infty}, \mcU_{k-1} \dist \distGam(\beta, \kappa_{s_{k}}). 
\]
For $0 \le i \le M, 1 \le j \le M$, let 
$\bar\pi_{i,j} \defined \pi_{ij}$ and for
$0 \le i \le M$, let 
$\bar\pi_{i,M+1} \defined 1 - \sum_{j=1}^{M} \pi_{ij}$. 
Integrating out $\{\kappa_{i}\}_{i \ge 1}$, the 
contribution to the likelihood from the \HGGP prior is
now
\[
\lefteqn{p(\mcU_{K}, \kappa_{0}, \bar\pi \given \beta,  \gamma_{0}, \gamma, \alpha_{0})} \\
&= p(\kappa_{0} \given \alpha_{0}, \gamma_{0}) 
   p(\bar\pi_{0} \given \alpha_{0}) 
   p(\bar\pi_{1:M} \given \beta\kappa_{0}\bar\pi_{0}) 
   p(\mcS_{K} \given \bar\pi_{1:M}) 
   p(\mcT_{K} \given \beta, \gamma, \kappa_{0}) \\
&\propto 
   \kappa_{0}^{\alpha_{0}-1}e^{-\gamma_{0}\kappa_{0}}
   \prod_{i=1}^{M} \distBeta\left(\frac{\bar\pi_{0i}}{1 - \sum_{j=1}^{i-1}\pi_{0,j}} \Bigg| 1, \alpha_{0}\right) \distDir(\bar\pi_{i} \given \beta\kappa_{0}\bar\pi_{0})
   \left(\prod_{k=1}^{K} \bar\pi_{s_{k-1},s_{k}}\right)
   p(\mcT_{K} \given \beta, \gamma, \kappa_{0}) \\
\begin{split}
&\propto 
  \kappa_{0}^{\alpha_{0}-1}e^{-\gamma_{0}\kappa_{0}}
   \prod_{i=1}^{M}\left\{\frac{\Gamma(1 + \alpha_{0})}{\Gamma(\alpha_{0})} \left(\frac{1 - \sum_{j=1}^{i}\pi_{0,j}}{1 - \sum_{j=1}^{i-1}\pi_{0,j}}\right)^{\alpha_{0} - 1}\Gamma(\beta\kappa_{0}) \prod_{j=1}^{M+1} \frac{\bar\pi_{ij}^{\beta\kappa_{0}\bar\pi_{0j}-1}}{\Gamma(\beta\kappa_{0}\bar\pi_{0j})}\right\} \\
&\phantom{\propto~} \times \prod_{k=1}^{K} \bar\pi_{s_{k-1},s_{k}}^{\beta\xi}
 \times \prod_{m=1}^{M} \frac{\Gamma(\beta(\kappa_0 + K_{m}))}{\Gamma(\beta)^{K_{m}}}
 \frac{(\prod_{j=1}^{K_{m}}t^{*}_{mj})^{\beta - 1}}{(\gamma + \sum_{j=1}^{K_{m}} t^{*}_{mj})^{\beta(\kappa_0 + K_{m})}}.
\end{split}
\]
We use a slightly different limiting process, with 
$\gamma_{0} = \kappa_{0} = \xi_{2}$, a positive constant, 
and scale the multinomial distributions \eqref{eq:hggp-state-dist}
by $\beta\xi$. 
Taking the logarithm and and ignoring $o(\beta)$ terms 
as before yields
\(
&\phantom{\sim~}\sum_{i=1}^{M}\left\{ \ln \alpha_{0} + \beta \kappa_{0}\ln \beta\kappa_{0} - \beta 
 + \sum_{j=1}^{M+1}\left\{ -\beta\kappa_{0}\bar\pi_{0,j} \ln(\beta\kappa_{0}\bar\pi_{0,j}) + \beta\kappa_{0}\bar\pi_{0,j} + \beta\kappa_{0}\bar\pi_{0,j}\ln \bar\pi_{ij}\right\}
\right\}  \\
&\phantom{\sim~} + \sum_{k=1}^{K} \beta \xi \ln \bar\pi_{s_{k-1},s_{k}} 
 + \sum_{m=1}^{M} \left\{ \textstyle\sum_{j=1}^{K_{m}}\beta \ln t^{*}_{mj} 
 - \beta K_{m}\ln \left([\gamma + t^{*}_{m\cdot}]/K_{m}\right) \right\} \\
&\sim 
   \sum_{i=1}^{M}\left\{ -\beta \xi_{1} 
    + \sum_{j=1}^{M+1}\left\{ -\beta\kappa_{0}\bar\pi_{0,j} \ln(\bar\pi_{0,j}) + \beta\kappa_{0}\bar\pi_{0,j}\ln \bar\pi_{ij}\right\}
\right\}  \\
&\phantom{\sim~}+ \sum_{k=1}^{K} \beta \xi \ln\bar\pi_{s_{k-1},s_{k}} 
 + \sum_{m=1}^{M} \left\{ \textstyle\sum_{j=1}^{K_{m}}\beta \ln t^{*}_{mj} 
 - \beta K_{m}\ln \left([\gamma + t^{*}_{m\cdot}]/K_{m}\right) \right\} \\
&\sim 
   - \beta\left\{ \xi_{1}M +  \xi\sum_{k=1}^{K} \ln\bar\pi_{s_{k-1},s_{k}} 
   + \sum_{m=1}^{M} \left\{\xi_{2}\kl{\bar\pi_{0}}{\bar\pi_{m}} - \textstyle\sum_{j=1}^{K_{m}} \ln t^{*}_{mj} - K_{m}\ln \left([\gamma + t^{*}_{m\cdot}]/K_{m}\right) \right\} \right\}.
\)
Thus, the objective function to minimize is
\[
\begin{split}
&\phantom{+}~\zeta\sum_{\ell=1}^{L} \ln\rho_{s_{\tau_{\ell}}x_{\ell}}
+ \xi \sum_{k=1}^{K} \ln\bar\pi_{s_{k-1},s_{k}}  + \xi_{1}M \\
&+ \sum_{m=1}^{M} \left\{\xi_{2}\kl{\bar\pi_{0}}{\bar\pi_{m}} - \textstyle\sum_{j=1}^{K_{m}} \ln t^{*}_{mj} - K_{m}\ln \left([\gamma + t^{*}_{m\cdot}]/K_{m}\right) \right\}.
\end{split}
\]

\section{Time-accuracy plots for the experiments}
\renewcommand\thefigure{\thesection.\arabic{figure}}  

\begin{figure}[t]
\vspace{-2mm}
\centering
\subfigure[]{%
	\includegraphics[trim = 0mm 0mm 20mm 0mm, clip, width = 0.32\textwidth]{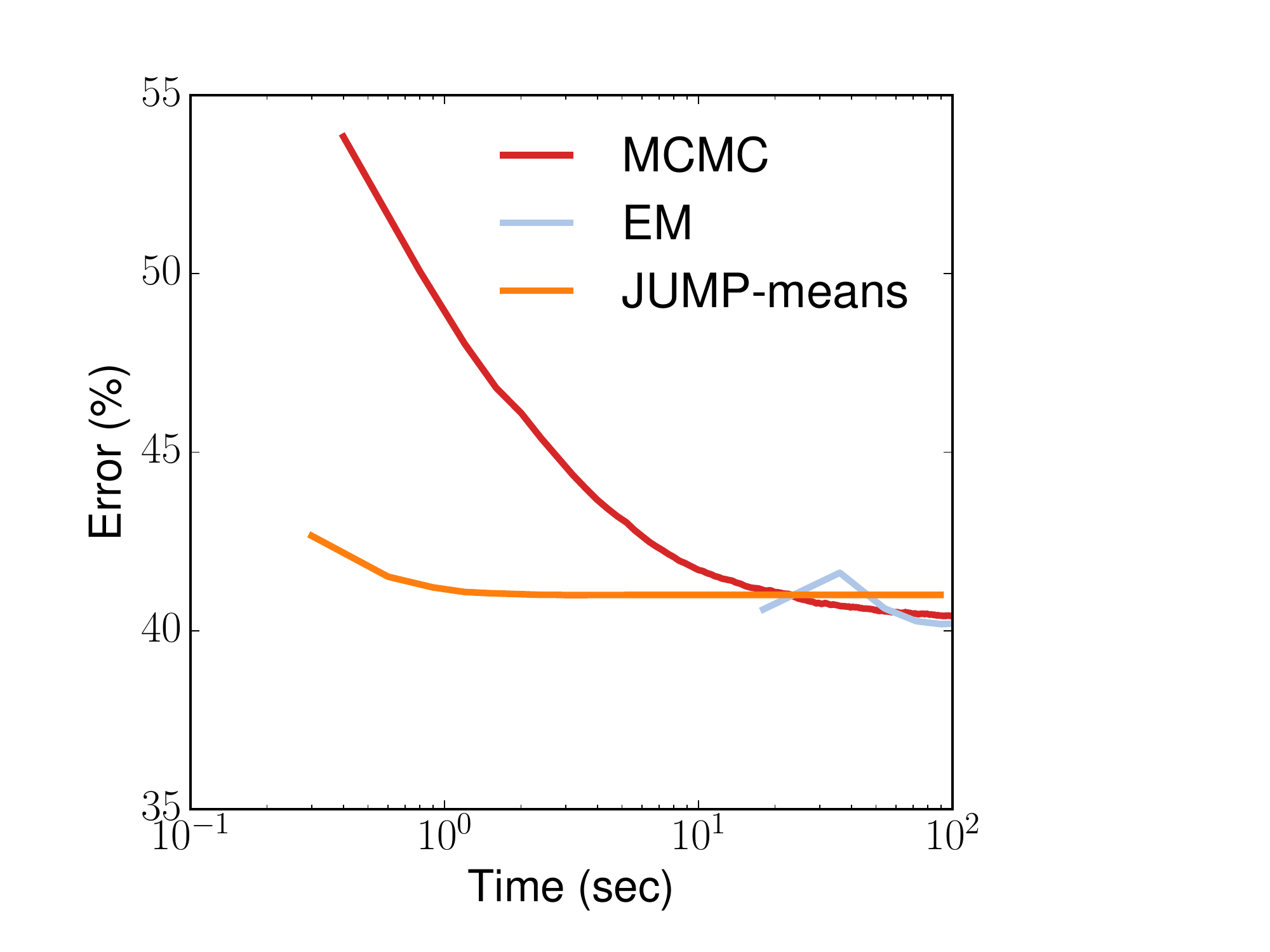}%
}
\subfigure[]{%
	\includegraphics[trim = 0mm 0mm 20mm 0mm, clip, width = 0.32\textwidth]{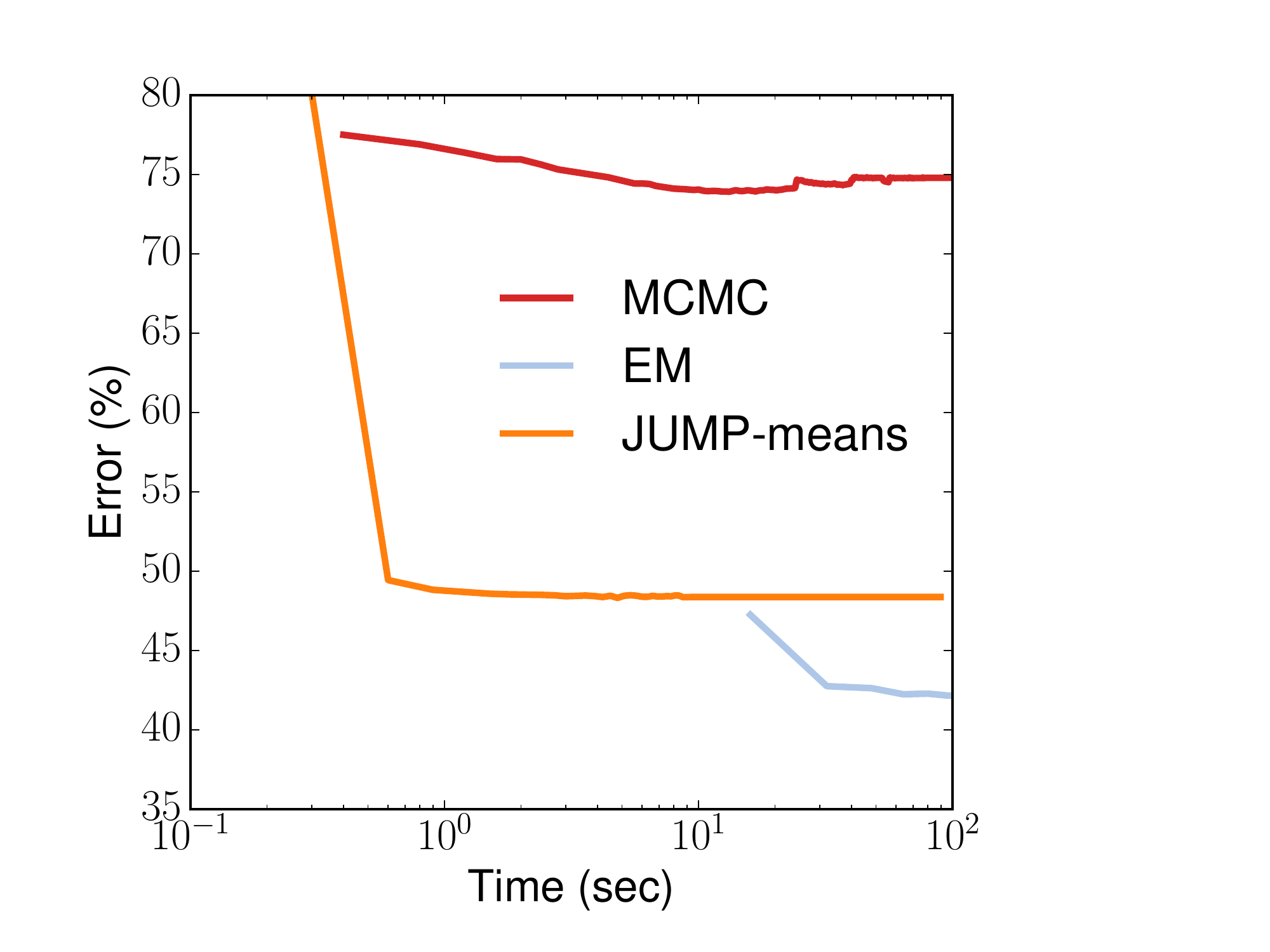}%
} \\
\subfigure[]{%
	\includegraphics[trim = 0mm 0mm 20mm 0mm, clip, width = 0.32\textwidth]{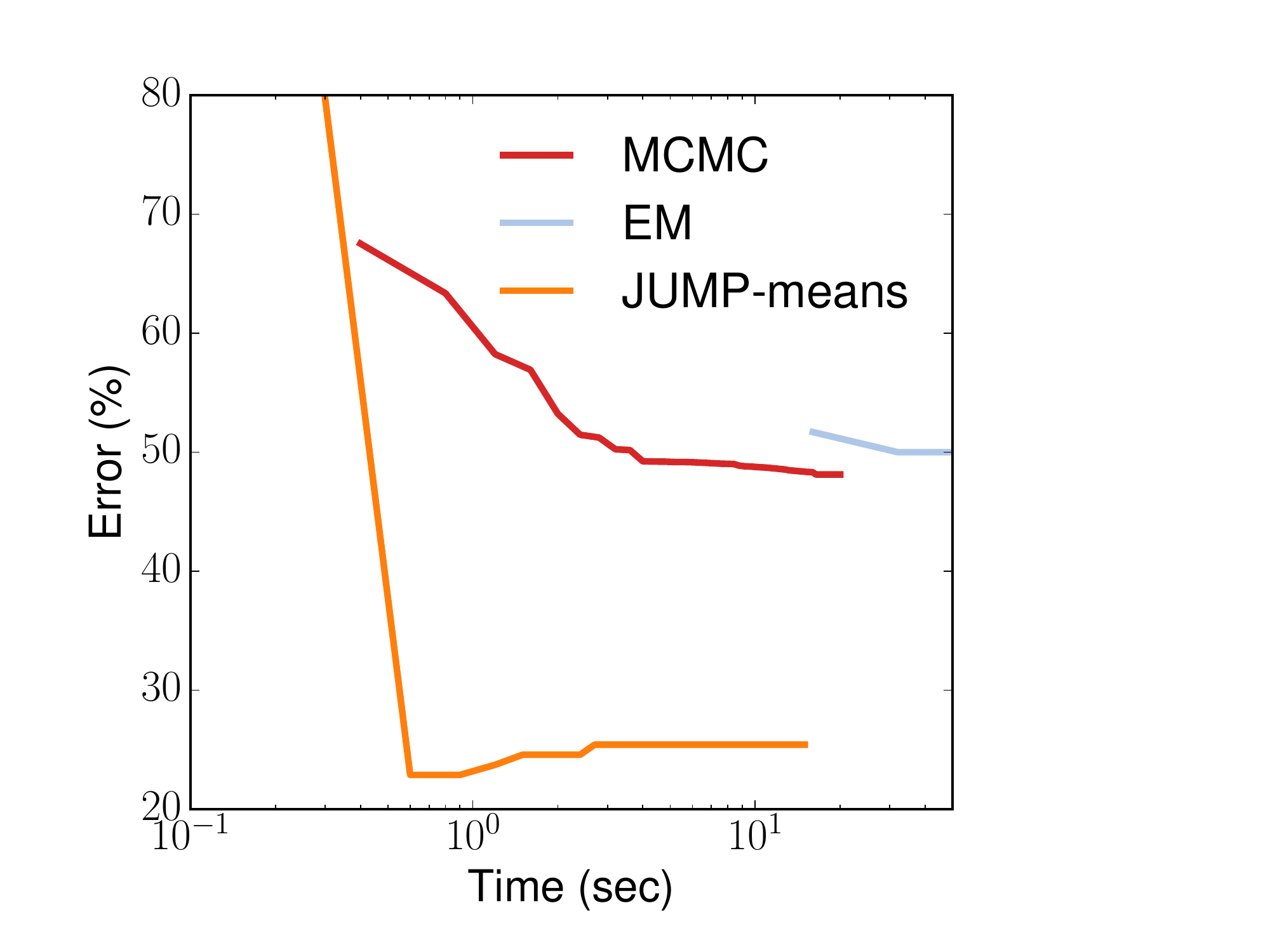}%
}
\subfigure[]{%
	\includegraphics[trim = 0mm 0mm 20mm 0mm, clip, width = 0.32\textwidth]{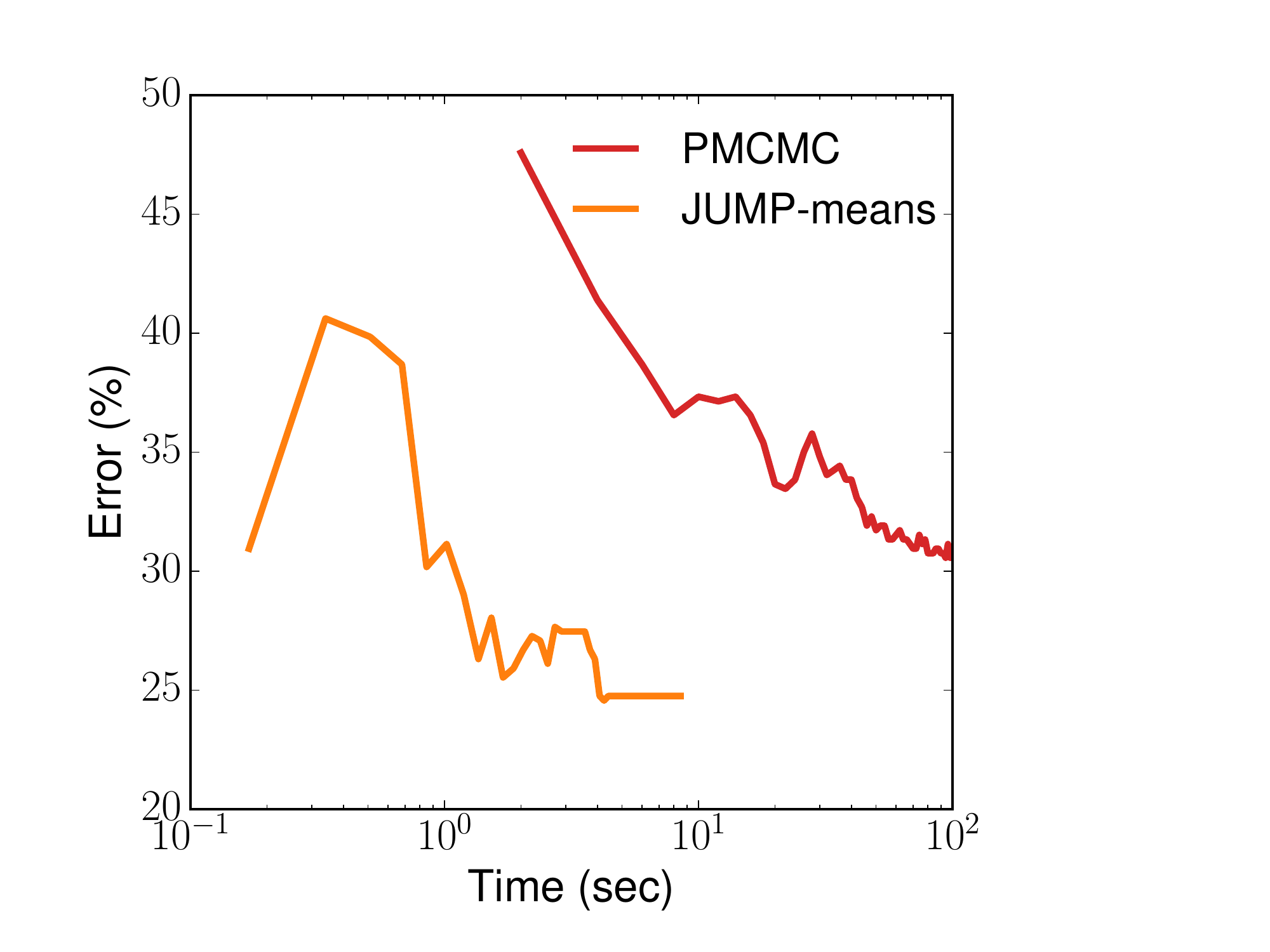}%
}
\vspace{-4mm}
\caption{\footnotesize{Mean error vs CPU runtime for 
\textbf{(a)} Synthetic 1; \textbf{(b)} Synthetic 2; \textbf{(c)} MS; and \textbf{(d)} MIMIC datasets.
In each case the JUMP-means algorithms have better or comparable performance to 
other standard methods of inference in MJPs.
}}
\label{fig:time-cpu}
\end{figure}

In the main paper we include the error versus iteration as it is more objective than time-accuracy results. In Fig. \ref{fig:time-cpu}, we compare the time-accuracy across different methods for different datasets. EM, PMCMC, and JUMP-means are implemented in Java and MCMC is implemented in Python. To plot the MCMC results, we give a speed boost of 100x in the results to compensate for Python's slow interpreter. From our experience with scientific computing applications, we believe this is a generous adjustment. Also we note that the EM implementation used in our experiments is not the most optimized in terms of time per iteration. However, our goal is to show that JUMP-means can achieve comparable performance with a reasonable implementation of MCMC and EM. 

\section{Scaling experiments}
For the scaling experiments we generated 4 datasets consisting of $10^2$ to $10^5$ sequences. All datasets are sampled from a single hidden state MJP with 5 hidden states and 5 possible observations. For the 20 observations in each sequence a Gaussian likelihood is used. Finally, for the held out results, we categorized the observations in 5 bins, removed $30\%$ of the data points and predicted their category.



\end{document}